\documentclass[lettersize,journal]{IEEEtran}
\usepackage{amsmath,amsfonts}
\usepackage[linesnumbered,ruled,vlined,boxed]{algorithm2e}
\usepackage{array}
\usepackage{subfig}
\usepackage{textcomp}
\usepackage{stfloats}
\usepackage{url}
\usepackage{verbatim}
\usepackage{graphicx}
\usepackage{cite}
\usepackage{textcomp}
\usepackage{stfloats}
\usepackage{url}
\usepackage{verbatim}
\usepackage{graphicx}
\usepackage{caption}
\usepackage{cite}
\usepackage{fancyhdr}
\usepackage{makeidx}
\usepackage{epsfig}
\usepackage{amsmath,amssymb,amsfonts}
\usepackage{multirow}
\usepackage{xcolor}
\usepackage{booktabs}
\usepackage{threeparttable}
\usepackage[normalem]{ulem}

\usepackage{verbatim}
\usepackage{color}
\usepackage{hyperref}
\usepackage{parskip}
\usepackage{algpseudocode}
\usepackage{amsthm}
\newtheorem{definition}{Definition}

\usepackage[capitalize]{cleveref}
\usepackage{tabularx}
\usepackage{booktabs} 
\usepackage{multirow} 
\begin{document}

\title{Compressing Deep Reinforcement Learning Networks with a Dynamic Structured Pruning Method for Autonomous Driving}

\author{Wensheng~Su,
        Zhenni~Li*,
        Minrui~Xu,
        Jiawen~Kang,\\
        Dusit~Niyato,~\IEEEmembership{Fellow,~IEEE,}
        and~Shengli~Xie,~\IEEEmembership{Fellow,~IEEE}
\thanks{This research is supported in part by the National Natural Science Foundation of China under Grants 62273106, 62203122, 62320106008, in part by the Guangdong Basic and Applied Basic Research Foundation 2023A1515011480, 2023A1515011159.~\textit{(Corresponding
author: Zhenni Li.)}}
\thanks{Wensheng Su is with the School of Automation, Guangdong University of Technology, Guangzhou 510006, China, and also with the Pillar of Information Systems Technology and Design, Singapore 487372 (e-mail: 2112004077@mail2.gdut.edu.cn).}
\thanks{Zhenni Li is with the School of Automation, Guangdong University of Technology, Guangzhou 510006, China, and also with the 111 Center for Intelligent Batch Manufacturing based on IoT Technology (GDUT), Guangzhou 510006, China (e-mail: lizhenni2012@gmail.com).}
\thanks{Minrui Xu and Dusit Niyato are with the School of Computer Science and Engineering, Nanyang Technological University, Singapore 639798 (e-mail: minrui001@e.ntu.edu.sg; dniyato@ntu.edu.sg).}
\thanks{Jiawen Kang is with the School of Automation, Guangdong University of Technology, Guangzhou 510006, China, and also with the Guangdong Key Laboratory of IoT Information Technology (GDUT), Guangzhou 510006, China (e-mail: kavinkang@gdut.edu.cn).}
\thanks{Shengli Xie is with the Key Laboratory of Intelligent Detection and The Internet of Things in Manufacturing (GDUT) and the Key Laboratory of Intelligent Information Processing and System Integration of IoT (GDUT), Ministry of Education, Guangzhou 510006, China (e-mail: shlxie@gdut.edu.cn).}}

\markboth{Journal of \LaTeX\ Class Files,~Vol.~14, No.~8, August~2021}%
{Shell \MakeLowercase{\textit{et al.}}: A Sample Article Using IEEEtran.cls for IEEE Journals}


\maketitle

\begin{abstract}
Deep reinforcement learning (DRL) has shown remarkable success in complex autonomous driving scenarios. However, DRL models inevitably bring high memory consumption and computation, which hinders their wide deployment in resource-limited autonomous driving devices. Structured Pruning has been recognized as a useful method to compress and accelerate DRL models, but it is still challenging to estimate the contribution of a parameter (i.e., neuron) to DRL models. In this paper, we introduce a novel dynamic structured pruning approach that gradually removes a DRL model's unimportant neurons during the training stage. Our method consists of two steps, i.e. training DRL models with a group sparse regularizer and removing unimportant neurons with a dynamic pruning threshold. To efficiently train the DRL model with a small number of important neurons, we employ a neuron-importance group sparse regularizer. In contrast to conventional regularizers, this regularizer imposes a penalty on redundant groups of neurons that do not significantly influence the output of the DRL model. Furthermore, we design a novel structured pruning strategy to dynamically determine the pruning threshold and gradually remove unimportant neurons with a binary mask. Therefore, our method can remove not only redundant groups of neurons of the DRL model but also achieve high and robust performance. Experimental results show that the proposed method is competitive with existing DRL pruning methods on discrete control environments (i.e., CartPole-v1 and LunarLander-v2) and MuJoCo continuous environments (i.e., Hopper-v3 and Walker2D-v3). Specifically, our method effectively compresses $93\%$ neurons and $96\%$ weights of the DRL model in four challenging DRL environments with slight accuracy degradation. 
\end{abstract}


\section{Introduction}\label{sec:Introduction}
\IEEEPARstart{R}{einforcement} learning (RL) has seen successful
applications in various domains including autonomous
driving~\cite{lin2023policy,kiran2021deep,zhu2021survey} and intelligent transportation systems (ITS)~\cite{haydari2020deep,zhao2020satopt,xu2022full}, it still faces challenges such as low-dimensional control commands and long-term decision making due to the various road geometry topology and real-time requirements of the autonomous vehicle. Different from the classical RL, deep reinforcement learning (DRL) algorithms adopt deep neural networks (DNN) with thousands of parameters to approximate the policy or value estimator of RL to solve complex and high-dimensional tasks, such as traffic management~\cite{christiano2017deep}, wireless sensing~\cite{HandFi_SenSys23,ji2022sifall,ji2021clnet} and vehicle control~\cite{mnih2013playing,nguyen2020deep}. However, training a DRL model demands heavy computational resources and storage. For instance, training a DRL model to solve the parallel task scheduling problem requires a large amount of data and a powerful computational system~\cite{han2016eie}. Furthermore, the storage of the trained model and the intermediate data generated from a dynamic and complicated autonomous driving environment also require a significant amount of storage space~\cite{yu2020deep,wang2021incorporating}. The heavy resource requirement hinders the application of DRL on resource-limited devices, such as embedded devices and FPGA, etc~\cite{li2021explainable,hazarika2022drl,wang2022joint}. Thus, it is critical to accelerate DRL model inference and reduce storage for resource-limited autonomous driving devices. For example, DRL models are used to make decisions about how to navigate and interact with the environment in the field of autonomous vehicles. If DRL models take up large storage space, it may be difficult to deploy them on a car onboard computer for real-time autonomous driving~\cite{wang2019autonomous}. 

To reduce the computation cost, recent works suggest that pruning technology is a promising approach for DRL model compression and acceleration. These pruning techniques can be broadly divided into two categories: non-structured pruning~\cite{livne2020pops,li2022compact,han2016eie,zhao2022double,zhao2023dynamic} and structured pruning~\cite{srivastava2014dropout,li2016pruning,wang2020weighted,roth2022non,zhang2022pointcutmix}. On the one hand, non-structured pruning methods directly prune weights of DRL models to obtain sparse weight matrices. Livne~\textit{et al.}~\cite{livne2020pops} apply non-structured pruning technology to DRL models for the first time and achieve a high compression ratio by using iterative hard threshold pruning. Because it is hard to determine an appropriate pruning threshold, Li~\textit{et al.}~\cite{li2022compact} propose an adaptive pruning technique with the $l_{1}$-norm sparsity regularizer, which can adaptively select important parameters of DRL models and obtain a high pruning ratio. However, there are two issues with existing non-structured pruning techniques: 1) It is challenging to utilize non-structured pruning methods for the acceleration of DRL training; 2) Non-structured pruning techniques often lead to irregular network structure, which is hard to implement for speedup on general hardware platforms~\cite{han2016eie}. 

To overcome these problems, structured pruning technologies focus on removing entire groups of consecutive parameters (such as weights or neurons) in a structured manner. Thus, the remaining network architecture can be used directly by the existing hardware. There are mainly two categories of methods for structured pruning. The first category is Dropout~\cite{srivastava2014dropout}, which aims at preventing a network from over-fitting by randomly dropping some neurons during training to obtain good performance. Moreover, it can not only help to improve the robustness and stability of models but also significantly reduce model overfitting. The second category is regularization-based methods, which add group regularization terms to the objective function and prune the weights by minimizing the objective function during training and improving the generalization of the model~\cite{li2016pruning,wang2020weighted,roth2022non,zhang2022pointcutmix}. However, there are several issues with existing structured pruning techniques: 1) Unlike supervised learning in which labels exist, it is hard to utilize them in the DRL domain; 2) Existing group regularization approaches tend to use a large and constant regularization factor for all weight groups in the network~\cite{wang2020weighted,zhang2022pointcutmix}, which has the following problems and cannot be directly applied in DRL domain~\cite{livne2020pops}; 3) Existing works apply a constant and static pruning threshold for pruning, which suggests that all weights in different groups are equally important, but it is difficult to determine an appropriate static threshold. 

\begin{figure}[t]
\centering
\includegraphics[width=0.5\textwidth]{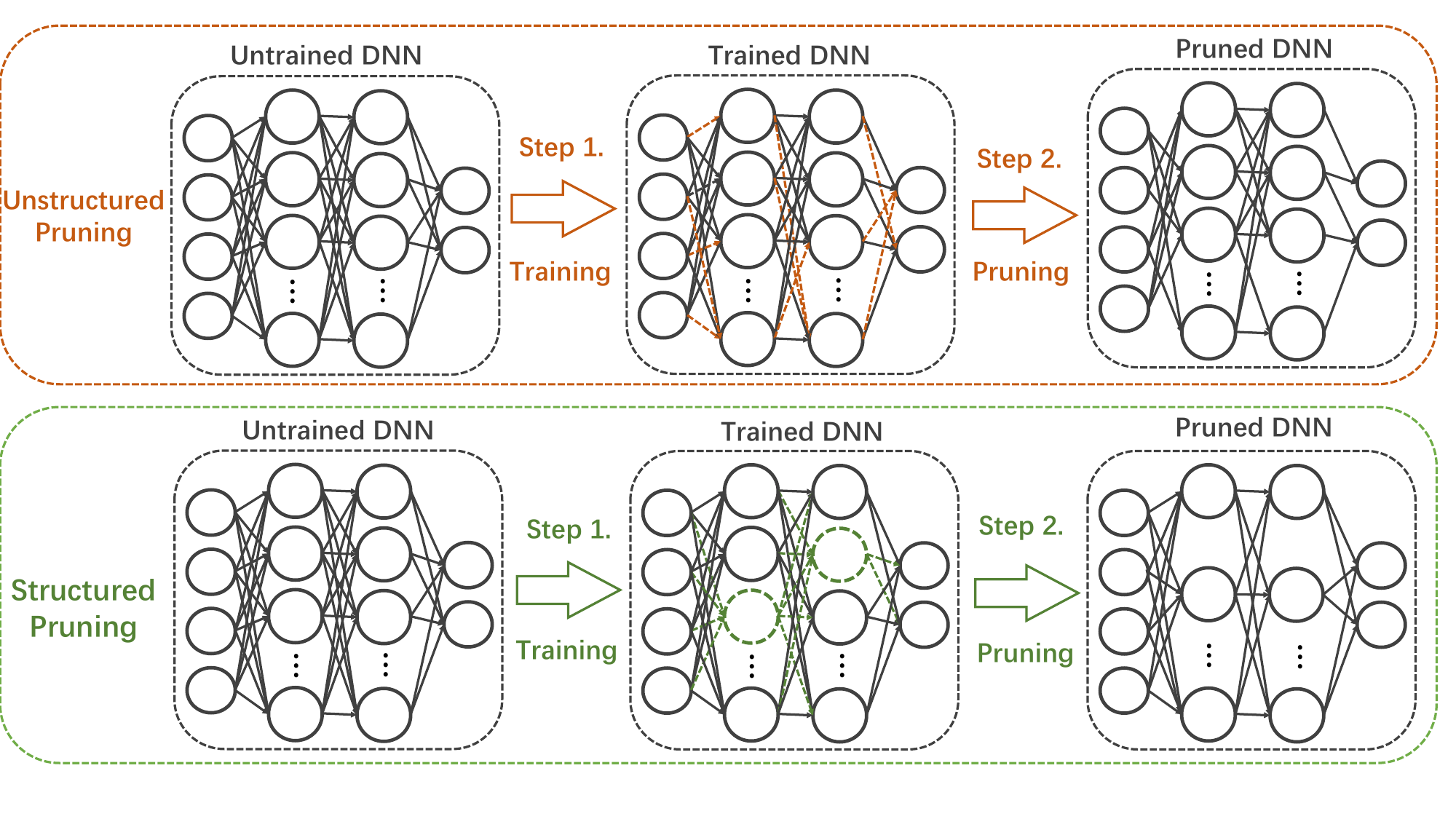}
\caption{An illustration of unstructured pruning versus structured pruning. Unstructured pruning results in nonstructured and irregular sparsity. Structured pruning targets at producing the structured and regular sparsity (dashed circles represent inactive neurons and dashed lines indicate removed weights).}
\label{fig:algorithm}
\end{figure}

In this paper, we propose a dynamic structured pruning method for deep reinforcement learning networks via a neuron-importance group sparse regularizer. Our method consists of two steps, i.e., training DRL models with a convex regularizer and removing unimportant neurons with a dynamic pruning threshold. To efficiently train the DRL model with a small number
of important neurons, we design a novel regularizer, i.e., neuron-importance group sparse regularizer, which imposes a penalty on redundant groups of neurons during training and encourages the selection of the most important neurons for model performance. When the importance of a neuron approaches the pruning threshold, the neuron is ready to be pruned. Different from pruning methods with a static threshold~\cite{han2015deep,li2022compact}, we design a novel structured pruning strategy to dynamically determine the pruning threshold, which is obtained by the number of training episodes and the coefficient of neuron importance. Moreover, we use a binary mask to remove unimportant neurons to create an efficient DRL model that is sparse (fewer parameters) and dense (fewer neurons). The whole training process does not require a pre-training model stage and a fine-tuning model stage, which improves the efficiency of the pruning. Therefore, our method can remove not only redundant and unimportant neurons from the DRL model but also achieve high and robust performance. We use discrete and continuous gym environments to validate the performance of the proposed method, which concludes a classic control environment (CartPole-v1), a box2D environment (LunarLander-v2), and MuJoCo continuous environments (Hopper-v3 and Walker2D-v3). Specifically, our method can outperform existing pruning technologies with less than 7\% neurons.

To the best of our knowledge, this is the first work to apply the structured pruning method based on regularization in the DRL domain. Our contributions to this paper are as follows:

\begin{itemize}
    \item To efficiently train the DRL model with a small number of important neurons, we propose to use a neuron-importance group sparse regularizer, which imposes a penalty on redundant groups of neurons that do not significantly impact the output of the DRL model. 
    \item To balance the trade-off between model performance and neuron sparsity, we propose a dynamic structured pruning strategy, which dynamically schedules the pruning threshold and gradually removes unimportant neurons according to the neuron importance coefficient and training episodes.
    \item Experimental results show that our training approach reduces 93\% neurons of the DRL model while achieving superior performance.
\end{itemize}

The remainder of the paper is organized as follows. The related work is presented in~\cref{sec:Related work}. The proposed algorithm and its optimization process are presented in~\cref{sec:The approach}. The experimental details and results are illustrated in~\cref{sec:Experiments}. Finally, we conclude the paper in~\cref{sec:Conclusion}.



\section{Related work}\label{sec:Related work}

Neural network pruning is the most widely used compression method for IoT devices, which can
transform the original dense deep neural network into a sparse network without a significant reduction in the network accuracy~\cite{wang2022paca,wu2020pruning,huang2022structured}. Pruning technologies can be divided into two categories, including unstructured pruning and structured pruning.
\subsection{Unstructured Pruning Techniques}
Unstructured pruning sets the elements of the weight matrix less than a threshold to zero, which is also known as sparse pruning. For example, Han~\textit{et al.}~\cite{han2015deep} proposed an iterative method to cut the network weight below a certain threshold. Yang~\textit{et al.}~\cite{yang2017designing} proposed a convolution neural network clipping algorithm, which cuts the network in a layer-wise manner by using energy consumption as the ranking criterion and restores network accuracy by using the least-squares method for local fine-tuning of each layer. However, these pruning methods~\cite{han2015deep,yang2017designing} can not directly apply in the DRL domain due to the unstable DRL training environment. To overcome this limitation, Livne~\textit{et al.}~\cite{livne2020pops} proposed iterative policy pruning with a hard threshold to obtain a sparse DRL agent. They also used knowledge distillation to fine-tune the performance of the network, because hard thresholds and unstable DRL training environments can cause accuracy loss effects on the DRL models. To adaptively select important parameters of DRL models, Li~\textit{et al.}~\cite{li2022compact} proposed an efficient pruning technique with low loss error to obtain a dense DRL model via $l_{1}$ sparse regularizer. However, there are still some problems with these unstructured pruning methods~\cite{livne2020pops,li2022compact}. First, the current computer hardware equipment cannot effectively train the non-compact network structure, so actually the running speed cannot be improved. Second, the pruned weights are irregularly distributed on the weight matrix. Therefore, directly adapting these methods to most deep reinforcement learning is not applicable, because reinforcement learning frameworks require frequent updates of neural network weights. Furthermore, this quantization only reduces memory usage without actual speedup, because the pruned weights are simply set to zero during computation. 
\subsection{Structured Pruning Techniques}
Structured pruning removes convolution kernels, filters, channels, and layers in the network and eliminates redundant connections in the network. Regularizers have been used to assess the correlation between convolution kernel and network performance and determine the filters to be pruned, such as $l_{1}$ regularizer~\cite{li2016pruning} and $l_{2}$ regularizer~\cite{he2018soft}. In~\cite{liu2017learning}, scaling coefficient normalization in the batch normalization layer was used to guide the thinning and pruning of the neural network structure. For example, Huang~\textit{et al.}~\cite{huang2021acceleration} accelerated neural networks based on an acceleration-aware finer-grained channel pruning method. Hu~\textit{et al.}~\cite{hu2018novel} proposed a channel clipping method for ultra-deep CNN compression, in which the trimming of the channel was formulated as a search problem solved by a genetic algorithm. He~\textit{et al.}~\cite{he2017channel} developed an effective channel clipping method based on LASSO regression and least-squares reconstruction, which accelerated VGG-16 by five times with a 0.3$\%$ error increase. Li~\textit{et al.}~\cite{li2021adaprune} proposed AdaPrune, which adaptively switches pruning between the input channel group and the output channel to accelerate deep neural networks. Most methods\cite{li2016pruning,he2018soft,he2017channel,li2021adaprune} require repeated experiments, which results in high costs of human and material resources. The most important problem of structured pruning is the determination of the redundant neuron in the network, while there is no clear criterion to solve this problem. Moreover, these methods only achieve good results in the image domain, and their performance in the unstable DRL training environment is unknown.

\section{The approach of dynamic Structured pruning for DRL networks}\label{sec:The approach}
In this section, we introduce our proposed method, i.e., dynamic structured pruning for compact deep reinforcement learning networks. We start from describing the Proximal Policy Optimization (PPO) algorithm, which is one of the state-of-the-art DRL methods. Second, we present the problem formulation of pruning redundant neurons in a DRL model training phase, where we propose to use a neuron-importance group sparse regularizer to train a DRL model and use binary masks to remove unimportant neurons with a dynamic pruning threshold.
\begin{table}[t]
\center
\caption{The symbol used in this paper}
\begin{tabular}{@{}l|l@{}}
\toprule 
Symbol                            & Definition \\ \midrule
$\mathcal{A}$                     & Action space \\
$\mathcal{R}$                     & Reward \\
$P$                   & Probability of transition \\
$\mathcal{S}$                     & State space \\
$h^{(l)}_{i}$                  & The $i$-th neuron in the $l$-th hidden layer         \\
$m_{i}^{l}$ & The mask of $i$-th neuron in the $l$-th layer \\
$p_i$,$p_f$ & Initial sparsity, target sparsity      \\
$N$ &   Total pruning steps \\
$O_{i}$           & The $i$-th neuron of the output layer \\
$\gamma$                          &  Discount factor \\
$\eta,\zeta$ & The learning rate of the gradient descent    \\
$\Omega^{(l)}_{i}$  &  Important neuron coefficience \\
$\|\cdot\|_0$ & Zero-norm\\
$\Delta$ & Pruning frequency\\
$\odot$ & Element-wise multiplication\\
\bottomrule
\end{tabular}
\end{table}

\subsection{Deep Reinforcement Learning Preliminaries}
In reinforcement learning, an agent interacts with an unknown environment to learn an optimal policy~\cite{jin2020reward}. The learning process is formulated as a Markov decision process (MDP) $\mathcal{M}=\langle\mathcal{S}, \mathcal{A}, \mathcal{R}, {P}, \gamma\rangle$, where $\mathcal{S}$ is the state space, $\mathcal{A}$ is the action space, $\mathcal{R}$ is the reward space, $P$ denotes the transition matrix, and $\gamma$ stands for the discount factor. Specifically, at the time step $t$, given the current state $\boldsymbol{s}_t \in \mathcal{S}$, the agent selects an action $\boldsymbol{a}_t \in \mathcal{A}$ according to the policy $\pi: \mathcal{S} \rightarrow \mathcal{A}$, then the agent receives a reward $r_t\in \mathcal{R}$ from the environment.

In the Proximal Policy Optimization (PPO) algorithm~\cite{schulman2017proximal}, the policy $\pi(\boldsymbol{s}_t, \boldsymbol{\boldsymbol{\theta}})$ is parameterized by an actor network with the weight parameter $\boldsymbol{\theta}$, and the state-value $V(\boldsymbol{s}_t, \boldsymbol{\theta})$ is parameterized by a critic network with the weight parameter $\boldsymbol{\theta}$. Specifically, the objective of the critic network is to minimize the TD error $\delta$, that is, 
\begin{equation}
    \delta=r_t+\gamma V\left(\boldsymbol{s}_t, \boldsymbol{w}\right)-V\left(\boldsymbol{s}_{t+1}, \boldsymbol{w}\right),
\end{equation}
where $V (\boldsymbol{s}_{t}, \boldsymbol{w})$ is the state-value of the current state $\boldsymbol{s}_{t}$, and $V (\boldsymbol{s}_{t+1}, \boldsymbol{w})$ is the state-value of the next state $\boldsymbol{s}_{t+1}$. Therefore, the loss function of the critic network is obtained by minimizing the expected value of the square of
the TD error $\delta$, i.e.,
\begin{equation}\label{critic loss}
    \min _{\boldsymbol{w}} F(\boldsymbol{w})=\min _{\boldsymbol{w}} \mathbb{E}\left[\left(r_t+\gamma V\left(\boldsymbol{s}_t, \boldsymbol{w}\right)-V\left(\boldsymbol{s}_{t+1}, \boldsymbol{w}\right)\right)^2\right],
\end{equation}
where $F(\boldsymbol{w})$ is the loss function of the critic network, $\mathbb{E}(\cdot)$ is the expectation function, $r_t$ is the reward at the time step $t$. Moreover, the objective of the actor network can be rewritten as follows:
\begin{equation}\label{actorloss}
    \max _{\boldsymbol{\theta}} J(\boldsymbol{\theta})=\max _{\boldsymbol{\theta}} \mathbb{E}\left[\min \left(\rho(\boldsymbol{\theta}) \hat{A}_t, g\left(\rho(\boldsymbol{\theta}), \epsilon_1, \epsilon_2\right) \hat{A}_t\right)\right],
\end{equation}
where $\rho_t(\boldsymbol{\theta})=\frac{\pi(s_t,\boldsymbol{\theta})}{\pi(s_t,\boldsymbol{\theta_{old}})}$, $\epsilon_1=1-\epsilon$, $\epsilon_2=1+\epsilon$ and $\epsilon$ is a hyperparameter. $g(\cdot)$ removes the incentive for moving $\rho(\boldsymbol{\theta})$ outside of the interval $\left[\epsilon_1, \epsilon_2\right]$ by clipping the important ratio. Here, $\hat{A}_t$ is an estimator of the advantage function at timestep $t$, which can be calculated as:
\begin{equation}
\hat{A}_t=-V\left(s_t\right)+\sum_{l=0}^{\infty} \gamma^l r_{t+l},
\end{equation}
where $\gamma$ is the discount factor and $r_{t+l}$ is the reward at $t+l$ time step.

\subsection{Notions and Problem Formulation}
In Proximal Policy Optimization (PPO) algorithm~\cite{schulman2017proximal}, the actor network and the critic network are deep neural networks (DNN) that are used to learn policies and value functions in DRL environments. In terms of the network structure, both the actor and critic networks are fully connected networks, which typically consist of an input layer, multiple hidden layers, and an output layer with a number of parameters, such as neurons and weights. Consider a $L$-layer actor network, where weights in the $l$-th fully connected layer are denoted by $\boldsymbol{\theta}^{(l)}$. $\boldsymbol{\theta}^{(l)}$ is a two-dimensional matrix and $1\le l \le L$. The actor and critic network receives state $\boldsymbol{s_t}$ at time step $t$. Here, we set the bias of the DNN to zero, and the input $\boldsymbol{s}_t$ enters the first layer and the output of the first layer is calculated by
\begin{equation}
\boldsymbol{h}^{(1)}=\sigma^{(1)}\left(\boldsymbol{\theta}^{(1)} \boldsymbol{s}_t \right),
\end{equation}
where $\sigma^{(1)}$ represents the nonlinear response of the first layer and is typically chosen to be the ReLU function~\cite{maas2013rectifier}. Therefore, the output of one layer is the input of the next, and the output of the $i$-th layer for $i=2, \ldots, N-1$ is given by
\begin{equation}
\boldsymbol{h}^{(l)}=\sigma^{(l)}\left(\boldsymbol{\theta}^{(l)} \boldsymbol{h}^{(l-1)}\right),
\end{equation}
where $\sigma^{(l)}$ represents the nonlinear response of the $l$ layer. The output of the actor network is action $\boldsymbol{a}_t$, which corresponds to the state $\boldsymbol{s}_t$ is
\begin{equation}
\boldsymbol{a}_t=\sigma^{(L)}\left(\boldsymbol{\theta}^{(L)} \boldsymbol{h}^{(L-1)}\right),
\end{equation}
where $\sigma^{(L)}$ represents the nonlinear response of the output layer. Our objective is to prune the redundant neurons and connected weights of the actor network, which are unimportant or useless for the
performance of the actor networks. Therefore, a binary mask $\boldsymbol{m}^{(l)}_i$ is introduced the pruning state of each neuron $o^{(l)}_{i}$, where $\boldsymbol{m}^{(l)}_i=1$ means the corresponding neuron should be reserved, and $\boldsymbol{m}^{(l)}_i=0$ means that the corresponding neuron should be pruned. At the beginning of training, the binary mask $\boldsymbol{m}^{(l)}_i$ is initialized all ones, because all neurons are initially considered important. As the model trains with the neuron-importance group sparse regularize, the binary mask $\boldsymbol{m}^{(l)}_i$ is updated based on the dynamic pruning threshold $\psi$ in~\cref{mask}. If the importance of a neuron is below the pruning threshold $\psi$, the corresponding binary mask $\boldsymbol{m}^{(l)}_i$ for the neuron is set to zero. Based on the binary mask $\boldsymbol{m}^{(l)}_i$, the output of the actor $\boldsymbol{a}_t$ is
\begin{equation}
    \boldsymbol{a}_t=\sigma^{(L)}\left(\boldsymbol{\theta}^{(L)} \boldsymbol{h}^{(L-1)}\odot \boldsymbol{m}^{(L)}\right),
\end{equation}

where $\odot$ indicates the element-wise multiplication of two matrices. Therefore, the loss function of the actor network is:

\begin{equation}
    J(\boldsymbol{\theta,m})=\mathbb{E}\left[\min \left(\rho(\boldsymbol{\theta,m}) \hat{A}_t, g\left(\rho(\boldsymbol{\theta,m}), \epsilon_1, \epsilon_2\right) \hat{A}_t\right)\right],
\end{equation}
where $\rho_t(\boldsymbol{\theta})=\frac{\pi(s_t,\boldsymbol{\theta})}{\pi(s_t,\boldsymbol{\theta_{old}})}$. Therefore, the loss function of actor network is:
\begin{equation}
\begin{aligned}
& \max _{\boldsymbol{\theta}} J(\boldsymbol{\theta,m})\\
& \text { s.t. }\sum_{l=1}^{L-1}\|\boldsymbol{m}^{(l)}\|_0 \leq C,
\end{aligned}
\end{equation}
where $J(\cdot)$ is a loss function (~\cref{actorloss}) of the actor network with the mask, $\|\cdot\|_0$ is zero-norm, $C$ is a hyperparameter that controls
the number of pruned neurons.~\cref{actorloss} can be transformed into the following form based Lagrange multiplier~\cite{bertsekas2014constrained}:
\begin{equation}\label{problem formulation}
    \max _{\boldsymbol{\theta}}J(\boldsymbol{\theta,m})-\alpha \sum_{l=1}^{L-1} \|\boldsymbol{m}^{(l)}\|_0,
\end{equation}
where $\alpha$ is a regularization coefficient and $\|\boldsymbol{\cdot}\|_0$ is the $l_0$-norm which represents the number of non-zero elements. Unfortunately, there is no efficient way to minimize the $l_0$-norm as it is
non-convex, NP-hard, and requires combinatorial search.

\begin{figure*}[!t]
\centering
\includegraphics[width=0.95\textwidth]{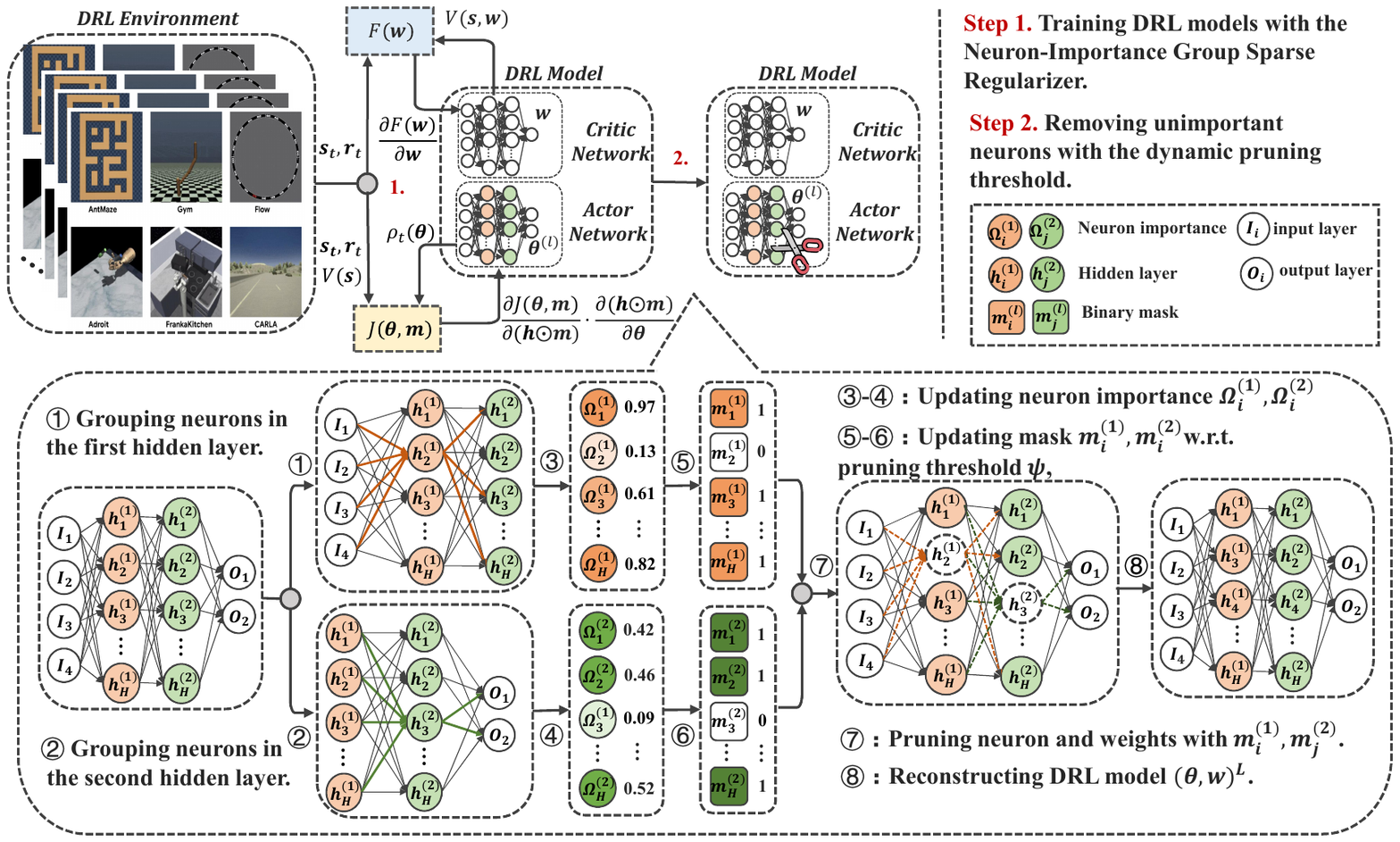}
\caption{An illustration of dynamic structured pruning algorithm for compact deep
reinforcement learning networks via
neuron-importance group sparse regularizer (dashed circles represent inactive neurons and dashed lines indicate removed weights).}
\label{fig:alg}
\end{figure*}

\subsection{Dynamic Structured Pruning with a Neuron-Importance Group Sparse Regularizer}

To address the previously mentioned non-convex challenges
in~\cref{problem formulation}, we propose our dynamic structured pruning with
a new convex regularization, i.e., neuron-importance group sparse regularizer. Specifically, the proposed algorithm is divided into two phases, which is illustrated in~\cref{fig:alg}. In the first phase, a convex regularizer is proposed to optimize the loss function of the actor network in a DRL model. In the second stage, the dynamic structured pruning method is proposed to train the compact and efficient actor network, which removes the least important neurons in a group.
\subsubsection{Training DRL model with a group sparse regularizer} Different from previous works~\cite{huang2022structured} that only consider
a neuron’s incoming or outgoing weights, we take both into
consideration. Our main objective is to obtain a sparse actor network with a significantly less number of parameters at both individual and group levels by using the proposed novel
combined regularizer, i.e., neuron importance regularization. We define the neuron importance of a network as follows: 
\begin{definition}[Neuron-Importance Group Sparse Regularizer]
Given a deep neural network $\boldsymbol{\theta}^{(L)}$ containing $L$ layers, the neuron importance $\Omega^{(l)}_{i}$ of the $i$-th neuron in the $l$-th layer can be computed as
\begin{equation}\label{def:neuron-imp}
    \Omega^{(l)}_{i}=\sum_j\left({\theta}_{i,j}^{(l)}\right)^2 \cdot \sum_k\left({\theta}_{k, i}^{(l+1)}\right)^2\cdot {m}^{(l)}_{i},
\end{equation}
where the sum of squares $\sum_j\left({\theta}_{i,j}^{(l)}\right)^2$ is the input weights of the neuron ${h}^{(l)}_{i}$ in $l$-th layer, and $m^{(l)}_{i}$ is the mask of the neuron ${h}^{(l)}_{i}$.
\end{definition}
Therefore, the~\cref{problem formulation} can be rewritten as follows:
\begin{equation}
\label{loss}
\begin{split}
    \max _{\boldsymbol{\theta}}\mathbb{E}\left[\min \left(\rho(\boldsymbol{\theta,m}) \hat{A}_t, g\left(\rho(\boldsymbol{\theta,m}), \epsilon_1, \epsilon_2\right) \hat{A}_t\right)\right]\\
-\lambda\sum_{i}\sum_{l}\Omega^{(l)}_{i}, 
\end{split}
\end{equation}
where $\lambda>0$ is the regularization coefficient. The first term is the loss function of the DNN, while the other regularization terms are convex. The object of the actor network is to maximize the expected value in~\cref{loss}. Moreover, the object of the critic network is obtained by minimizing the expected value of the square of
the TD error as~\cref{critic loss}. Since the cost
functions of both the actor network and the critic network are
smooth and convex, the policy gradient algorithm~\cite{silver2014deterministic} is used
to update the parameters $\boldsymbol{\theta}$ and $\boldsymbol{w}$. Therefore, the parameters $\boldsymbol{\theta}$ and $\boldsymbol{w}$ can be iteratively updated via the stochastic gradient ascent as

\begin{equation}
\boldsymbol{\theta}^{(l)}\gets \boldsymbol{\theta}^{(l)}-\eta\frac{\partial J({\boldsymbol{\theta}})}{\partial(\boldsymbol{h}^{(l)}\odot \boldsymbol{m}^{(l)})}\cdot\frac{\partial (\boldsymbol{h}^{(l)}\odot \boldsymbol{m}^{(l)})}{\partial\boldsymbol{\theta}^{(l)}},
\label{actor weight}
\end{equation}
\begin{equation}
    \boldsymbol{w}^{(l)} \leftarrow \boldsymbol{w}^{(l)}-\zeta \frac{\partial F({\boldsymbol{w}})}{\partial\boldsymbol{w}^{(l)}},
\label{critic weight}
\end{equation}
where $\eta$ and $\zeta$ are learning rates of the gradient descent, $\odot$ is element-wise multiplication. Therefore, a sparse actor network is achieved by solving~\cref{actor weight}, and the neuron-importance group sparse regularizer pushes the coefficients of unimportant neurons close to zero. 

\subsubsection{Removing unimportant neurons via dynamic pruning threshold}
To remove unimportant neurons, structured pruning is a promising technique to reduce the size and complexity of a model by removing redundant neurons or weights. Dynamic structured pruning of unimportant neurons contains two steps. We first calculate the pruning threshold, then update the binary mask for pruning. The pruning threshold is an important factor determining which parameters or connections should be kept and which should be removed during the pruning process. However, finding the optimal pruning threshold requires careful tuning and evaluation, which consists of two main challenges. First, the search space of threshold values is complex and computationally intractable for exhaustive exploration. Second, simply setting the threshold as a one-time static value could negatively affect the model accuracy~\cite{ham20203,wang2021spatten}. To
mitigate these challenges, we propose to learn the dynamic threshold value with the object of gradually increasing model sparsity as the number of iterations increases. Therefore, we define the dynamic pruning threshold as follows:

\begin{definition}[Dynamic Pruning Threshold]
Given a deep neural network $\boldsymbol{\theta}^{(L)}$ containing $L$ layers, the pruning threshold $\psi_{t}$ can be computed by
\begin{equation}
\psi_{t}=\sum_{i}\sum_{l}\Omega^{(l)}_{i}\cdot p_{t},
\label{Dynamic Pruning Threshold}
\end{equation}
\begin{equation}
p_{t}=p_f+\left(p_i-p_f\right)\left(1-\frac{t-t_i}{N \Delta t}\right)^3.
\end{equation}
where $p_i$ is the initial sparsity, $p_f$ is the target sparsity, $t_0$ is the starting epoch of gradual pruning, $N$ is the total pruning steps.
\end{definition}
After computing the value of the dynamic pruning threshold, we sort the importance of neurons in order from small to large and apply the mask to the neuron that is under the threshold. The mask is updated by:
\begin{equation}
    m_i^{(l)}= 
    \begin{cases}$1$, & \text{if abs}\left(m_i^{(l)} \theta^{(l)}_i\right) \geq \psi \\ 
    0, & \text{if abs}\left(m_i^{(l)} \theta^{(l)}_i\right)<\psi\end{cases}
\label{mask}
\end{equation}
where $\psi$ is the dynamic pruning threshold and $\text{abs}(\cdot)$ is the absolute value. After applying masks to unimportant neurons, only a small proportion of neurons remain. Then, we construct a compact actor network according to the redundancy of the sparse actor network. The procedure of the dynamic structured pruning is illustrated in~\cref{alg:1}.

\subsection{Complexity Analysis}
In~\cref{alg:1}, we consider that an actor network in the DRL model is a fully connected deep neural network (DNN) with $L$ layers. The process of~\cref{alg:1} includes training DRL models with
a group sparse regularizer (\cref{Alg:Compute actor-critic loss,Alg:Updating the actor,Alg:Compute neuron importance}) and removing unimportant neurons
with a dynamic pruning threshold (\cref{Alg:Compute the dynamic pruning threshold,Alg:Updating the mask,Alg:Remove neurons,Alg:Remove parameters,Alg:pruning condition}). The number of multiplications during the forward propagation and backward propagation is equal, which is given by
\begin{equation}
    D=S \cdot h^{(1)}+\sum_{l=1}^{L-1} h^{(l)} h^{(l+1)},
\end{equation}
where $S$ is the size of the state vector and $h^{(l)}$ is the number
of neurons in the $l$-th layer. Therefore, the computational complexity of the forward and backpropagation for one episode is $O(D)$. During the process of removing unimportant neurons with masks (\cref{Alg:Compute the dynamic pruning threshold,Alg:Remove neurons}), we define the computational complexity around $O(\sum^{L-1}_{l=1}h^{(l)})$ for updating every neuron during training the DRL model. Finally, the total computational complexity on $T$ episodes is $O(TS)+O(T\sum^{L-1}_{l=1}h^{(l)})$.

\begin{algorithm}
\KwIn{A DRL training environment $Env$.}
\KwOut{A compact DRL model $(\boldsymbol{\theta},\boldsymbol{w})^{(L)}$.}
Initialize a DRL model $(\boldsymbol{\theta},\boldsymbol{w})^{(L)}$, training episodes $T$, reward $r$, binary mask $\boldsymbol{m}$. \\
\For{$t=1,\cdots,T$}{
    The DRL model $(\boldsymbol{\theta},\boldsymbol{w})^{(L)}$ interacts with $Env$.\label{Alg:interact}\\
    Compute neuron importance $\Omega^{(l)}_{i}$ by~\cref{def:neuron-imp}.\label{Alg:Compute neuron importance}\\
    Compute $J(\boldsymbol{\theta})$, $L(\boldsymbol{w})$ by~\cref{loss,critic loss}.\label{Alg:Compute actor-critic loss}\\
    Updating the actor network parameter $\boldsymbol{\theta}^{(l)}$ by~\cref{actor weight}.\label{Alg:Updating the actor}\\
    Updating the critic network parameter $\boldsymbol{w}^{(l)}$ by~\cref{critic weight}.\label{Alg:Updating the critic}\\
    Compute the dynamic pruning threshold $\psi$ for the actor network $\boldsymbol{\theta}^{(L)}$ by~\cref{Dynamic Pruning Threshold}.\label{Alg:Compute the dynamic pruning threshold}\\
    Updating the mask $m_i^{(l)}$ for the actor network $\boldsymbol{\theta}^{(L)}$ by~\cref{mask}.\label{Alg:Updating the mask}\\
\If{$\Omega^{(l)}_{i}<\psi$\label{Alg:pruning condition}}{
    Remove the $i$-th neuron in $l$-th layer from the actor network $\boldsymbol{\theta}^{(L)}$ with the mask $m_i^{(l)}$.\label{Alg:Remove neurons}\\
    Remove parameters $\boldsymbol{\theta}$ connected to the removed neuron.}\label{Alg:Remove parameters}
}
Reconstruct a compact DRL model $(\boldsymbol{\theta},\boldsymbol{w})^{(L)}$ with the parameter of the actor network $\boldsymbol{\theta}^{(L)}$ and critic network $\boldsymbol{w}^{(L)}$.
\caption{Dynamic Structured Pruning for Deep Reinforcement Learning Networks\label{alg:1}}
\end{algorithm}
\section{Experiments}\label{sec:Experiments}
In this section, we evaluate the performance of the proposed dynamic structured pruning in deep reinforcement learning. First, we introduce the details of the experiment settings. Second, we illustrate how parameter choices affect performances in sparsity and rewards. Third, we compare baselines with corresponding optimal parameters. Finally, we visualize the model structure and the number of neurons to indicate the sparsity and performance of the models.

\subsection{Experiment settings}
To evaluate the performance of the proposed dynamic structured pruning in deep reinforcement learning, the experiments are evaluated on two discrete environments (Cart Pole and Lunar Lander) and two MuJoCo continuous control tasks (Hopper and Walker2D). The evaluation is carried out on a server with a 24-core Intel Xeon Platinum 6164 CPU and 32GB DDR4-2400 memory.

\subsubsection{Environment details}
Since the proposed algorithm is based on the PPO architecture, we evaluate our proposed approach on four challenging environments (CartPole-v1\footnote{\url{https://www.gymlibrary.dev/environments/classic_control/cart_pole}}, LunarLander-v2\footnote{\url{https://www.gymlibrary.dev/environments/box2d/lunar_lander}}, Hopper-v2\footnote{\url{https://www.gymlibrary.dev/environments/mujoco/hopper}}, and Walker2d-v2\footnote{\url{https://www.gymlibrary.dev/environments/mujoco/walker2d}}): 1) CartPole-v1. CartPole is a classic control problem in which an agent controls a cart that is attached to a pole by a hinge. In a discrete version of the CartPole problem, the action space is discrete, and the goal of the CartPole task is to balance the pole vertically on the cart for at least 195 timesteps. 2) LunarLander-v2. In a discrete LunarLander control problem, the agent is the lunar lander spacecraft, and the goal is for the agent to learn to navigate and land on the surface of the moon using a set of discrete actions, which is simulated using the Box2D physics engine. The action space of Lunar Lander consists of four discrete actions: firing the left thruster, firing the main thruster, firing the right thruster, and doing nothing. The state space includes the position, velocity, and orientation of the spacecraft, as well as the fuel remaining in the thrusters. 3) Hopper-v3. The Hopper is a Mujoco continuous control task in which a one-legged robot learns to hop forward as efficiently as possible. The state space of the Hopper problem concludes with the position, velocity, and acceleration of the robot's body. The action space consists of the force applied by the robot's leg to the ground. 4) Walker2d-v3. The Walker2d environment simulates a two-legged robot with the MuJoCo physics engine. In the Walker2d environment, the state space consists of the positions, velocities, and accelerations of the various joints and bodies of the robot, as well as the forces and torques acting on them. The action space consists of the desired joint torques applied to the robot's legs, which can be used to control its movement. The goal of the Walker2d environment is to get the robot to walk forward as far as possible without falling over.

\subsubsection{Settings of baselines}
To validate the performance of our proposed algorithm, we compared it with the state-of-the-art pruning methods based on the PPO architecture.
\begin{itemize}
    \item \textit{PPO-SSL}~\cite{wen2016learning}: Structured Sparsity Learning (SSL) method is a regularization method, which applies group lasso as the regularizer for deep neural networks (DNNs) to encourage sparsity in the model weights. It applies a penalty to the sum of the absolute values of the weights in each group to reduce redundant weights in DNNs. We search the coefficient of the penalty $\lambda\in[10^{-3},10^{-8}]$, and the pruning rate $r\in\{80\%, 85\%, 90\%, 93\%\}$.
    \item \textit{PPO-Dropout}~\cite{chen2021adaptive}: Dropout is a regularization technique, which randomly sets a percentage of neurons in a layer to zero to prevent DNNs overfitting. We search the dropout ratio $r\in\{80\%, 85\%, 90\%, 93\%\}$.
    \item \textit{PPO-$l_1$ Lasso}~\cite{li2022compact}: The $l_1$ lasso is a method for reducing the complexity of the DNNs by adding a penalty term of the absolute value of the weight to the loss function, which helps to promote sparsity in DNNs by driving the unimportant weights to zero. We search the coefficient of the penalty $\lambda\in[10^{-3},10^{-8}]$.
    \item \textit{PPO-PoPS}~\cite{livne2020pops}: Policy Pruning and Shrinking (PoPS) is a technique for reducing the complexity of deep reinforcement learning (DRL) models with a static pruning threshold, which removes the unimportant weights below the threshold. We search the pruning ratio $r\in\{80\%, 85\%, 90\%, 93\%\}$.
    \item \textit{PPO-SGS}~\cite{aghdaie2022morph}: The Structured Group Sparsity (SGS) is a technique that applies group $l_1$ sparsity constraints on the first fully connected layer of the deep neural network, thereby selecting the most discriminative wavelet sub-bands. This approach reduces the dimensionality of the input data and improves classification performance. In order to compare the different methods under the same parameters, we search the coefficient of the penalty $\lambda\in[10^{-3},10^{-8}]$, and the pruning rate $r\in\{80\%, 85\%, 90\%, 93\%\}$.
\end{itemize}

\subsubsection{Metrics}
The details of the metrics used for evaluating the performance of the proposed approach are discussed as follows.
\begin{itemize}
    \item \textit{Reward}: The standard metric used in the DRL domain to measure an agent's performance. The return $R$ is the sum of rewards in an episode with $T$ steps, which is calculated as follows:  
    \begin{equation}
        R=\sum_{t=1}^T r_t
    \end{equation}
    where the $r_t$ is a reward obtained in time step $t$.
    \item \textit{Model Parameters (\#Weights)}: The total number of parameters in the model.
    Considering a sparse network with $L$ fully connected layers, we calculate the number of weights of the model $M_w$ as:
    \begin{equation}
        M_w=\sum_{l=1}^L\left(1-S^{(l)} \right) I^{(l)} O^{(l)}
    \end{equation}
    where $S^{(l)}$ is the sparsity, $I^{(l)}$ is the input dimensionality, and $O^{(l)}$ is the output dimensionality of the $l$-th layer.
    \item \textit{Model Size (\#Neurons)}: The total number of neurons in the model. The size of a model $M_n$ is typically calculated by adding up the total number of neurons:
    \begin{equation}
        M_n=\sum_{l=1}^{L-1}\|\boldsymbol{m}^{(l)}\|_0
    \end{equation}
    where $\boldsymbol{m}^{(l)}$ is the mask used in layer $l$, $\|\cdot\|_0$ is the $L_0$ norm, and $L$ is the number of layers in the model.
     \item \textit{FLOPs}: Float point operations per second (FLOPs) is a criterion to quantify the computational complexity of a model. Consider a sparse network with $L$ fully connected layer, the required FLOPs for a forward pass is computed as follows:
\begin{equation}
    \text { FLOPs }=\sum_{l=1}^L\left(1-S^{(l)}\right)\left(2 I^{(l)}-1\right) O^{(l)},
\end{equation}
where $S^{(l)}$ is the sparsity, $I^{(l)}$ is the input dimensionality, and $O^{(l)}$ is the output dimensionality of the $l$-th layer. 
\end{itemize}

\subsection{Hyperparameters and Sensitivity analysis}

\begin{figure*}[!t]
\centering
\subfloat[]{\includegraphics[width=1.78in]{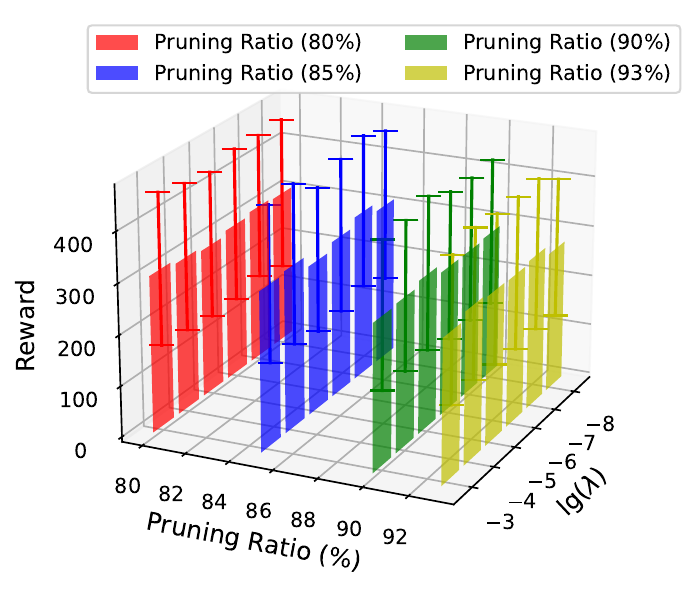}}
\hfil
\subfloat[]{\includegraphics[width=1.78in]{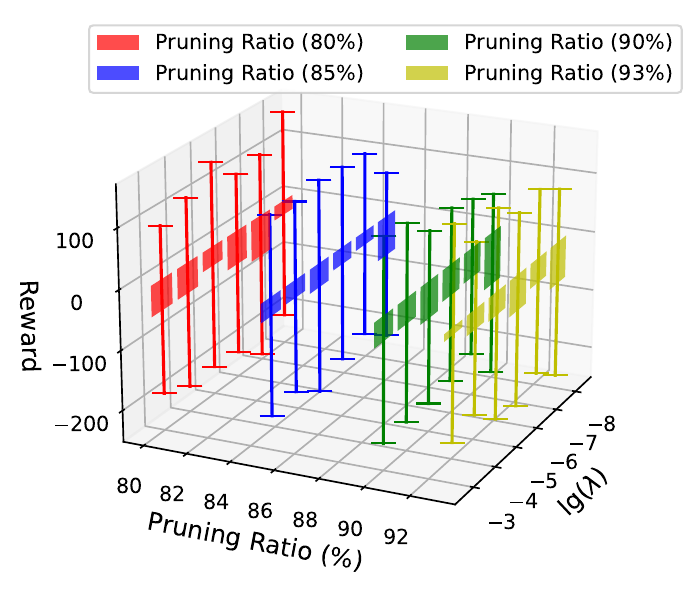}}
\hfil
\subfloat[]{\includegraphics[width=1.78in]{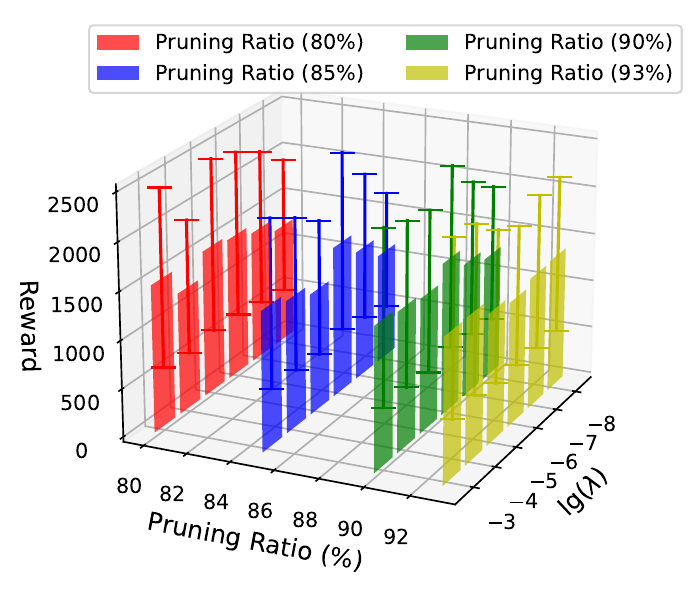}}
\hfil
\subfloat[]{\includegraphics[width=1.78in]{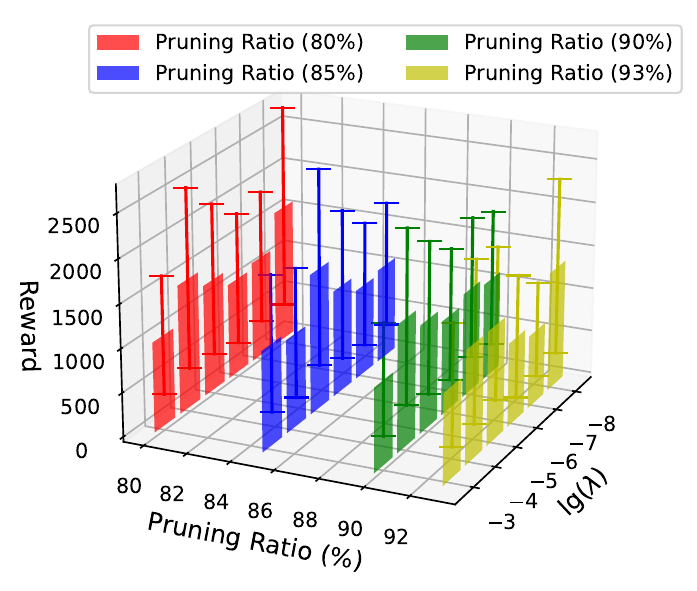}}
\caption{Performance results of the PPO-DSP method in Cart Pole, Lunar Lander, Hopper, and Walker environments with different settings of $\lambda$ and pruning ratio $r$. (a) Cart Pole. (b) Lunar Lander. (c) Hopper. (d) Walker2D.}
\label{fig:E1-con}
\end{figure*} 

\begin{figure*}[!t]
\centering
\subfloat[]{\includegraphics[width=1.78in]{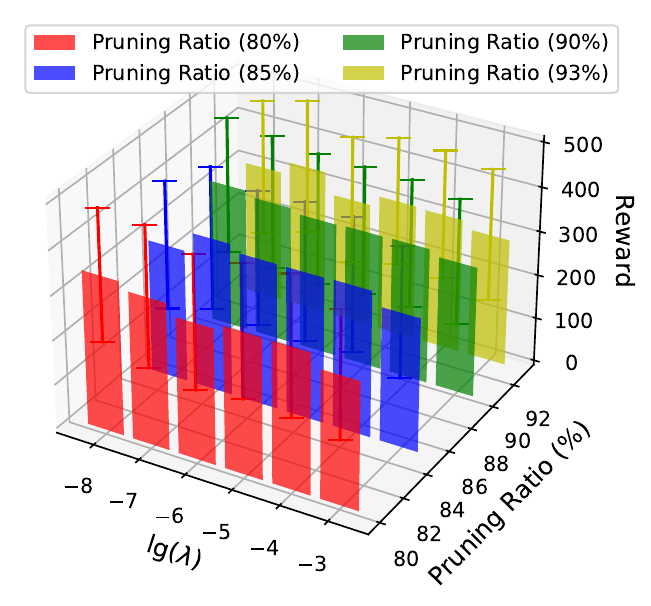}}
\hfil
\subfloat[]{\includegraphics[width=1.78in]{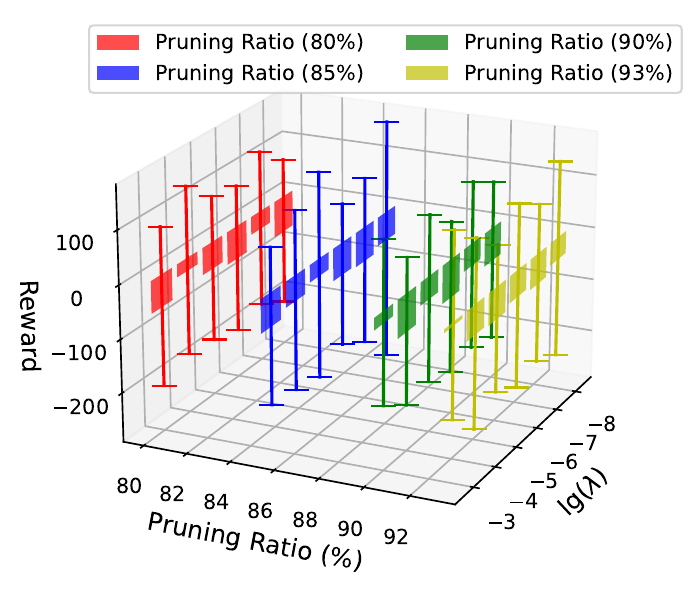}}
\hfil
\subfloat[]{\includegraphics[width=1.78in]{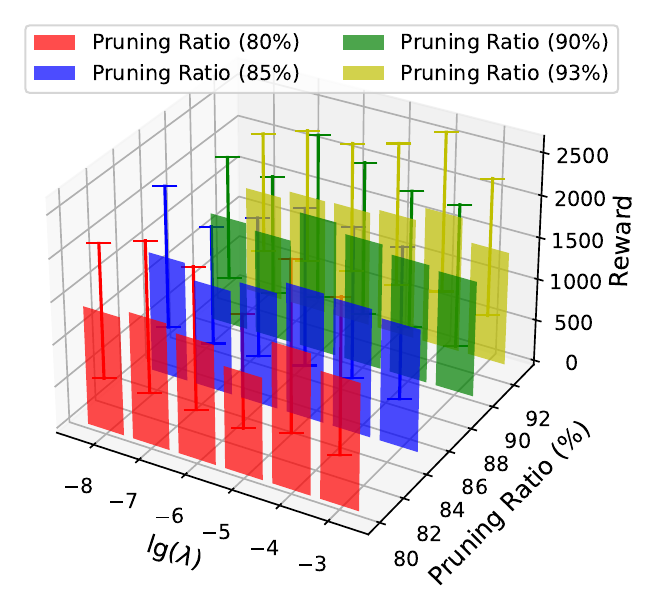}}
\hfil
\subfloat[]{\includegraphics[width=1.78in]{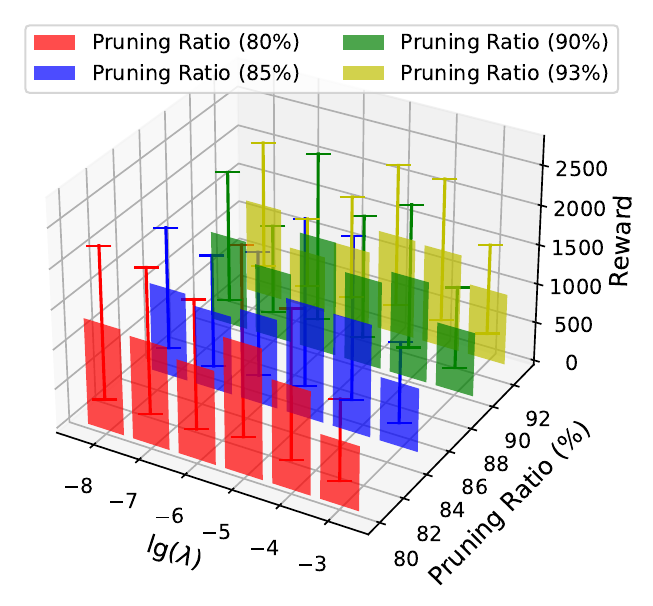}}
\caption{Performance results of the PPO-SSL method in Cart Pole, Lunar Lander, Hopper, and Walker environments with different settings of $\lambda$ and pruning ratio $r$. (a) Cart Pole. (b) Lunar Lander. (c) Hopper. (d) Walker2D.}
\label{fig:E1-glasso}
\end{figure*}
\begin{figure*}[!t]
\centering
\subfloat[]{\includegraphics[width=1.78in]{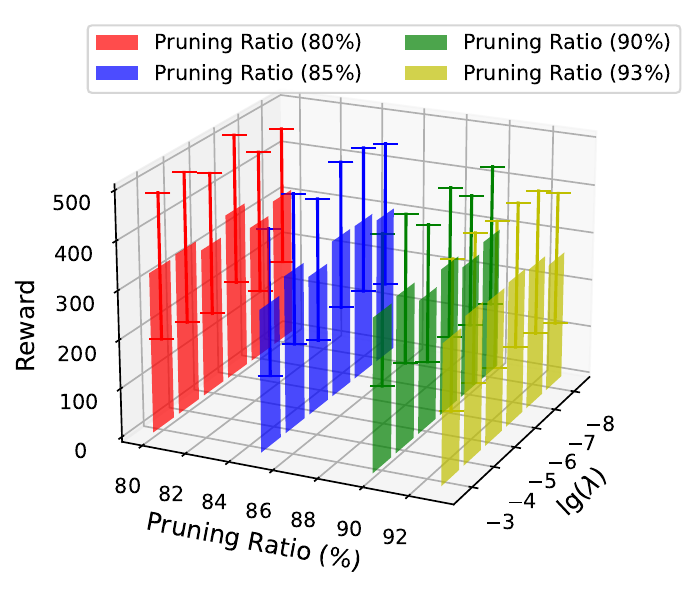}}
\hfil
\subfloat[]{\includegraphics[width=1.78in]{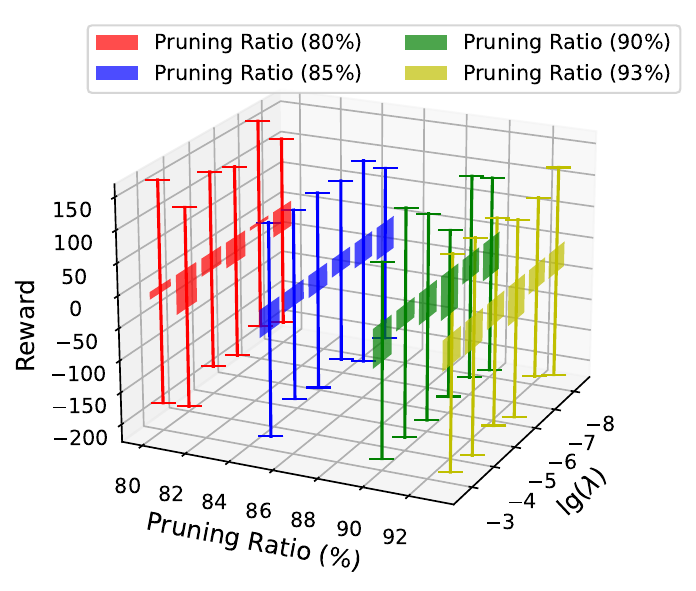}}
\hfil
\subfloat[]{\includegraphics[width=1.78in]{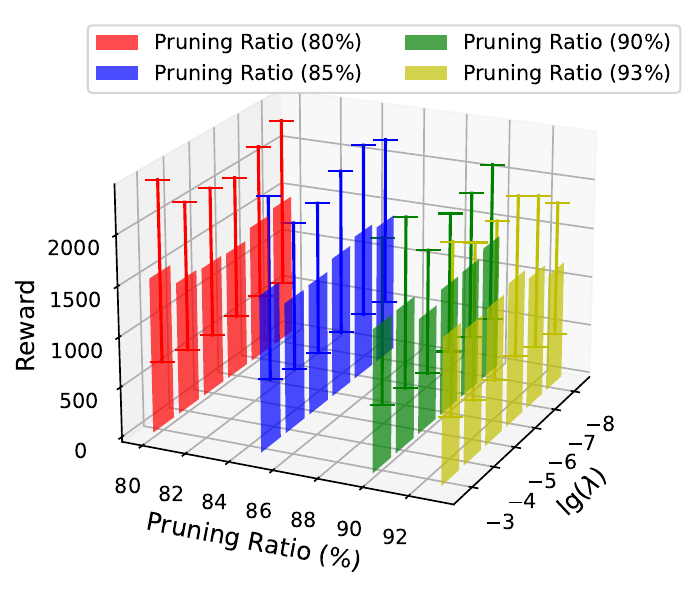}}
\hfil
\subfloat[]{\includegraphics[width=1.78in]{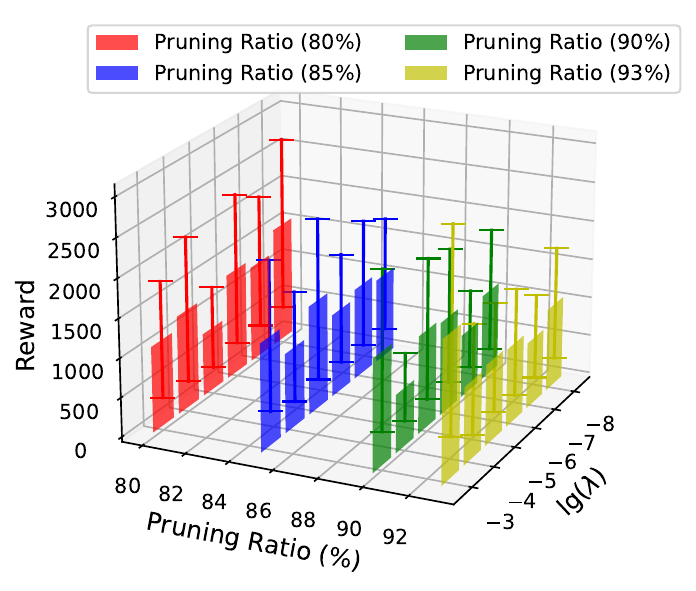}}
\caption{Performance results of the PPO-SGS method in Cart Pole, Lunar Lander, Hopper, and Walker environments with different settings of $\lambda$ and pruning ratio $r$. (a) Cart Pole. (b) Lunar Lander. (c) Hopper. (d) Walker2D.}
\label{fig:E1-l21norm}
\end{figure*} 
\begin{figure*}[!t]
\begin{center}
\subfloat[]{\includegraphics[width=1.8in]{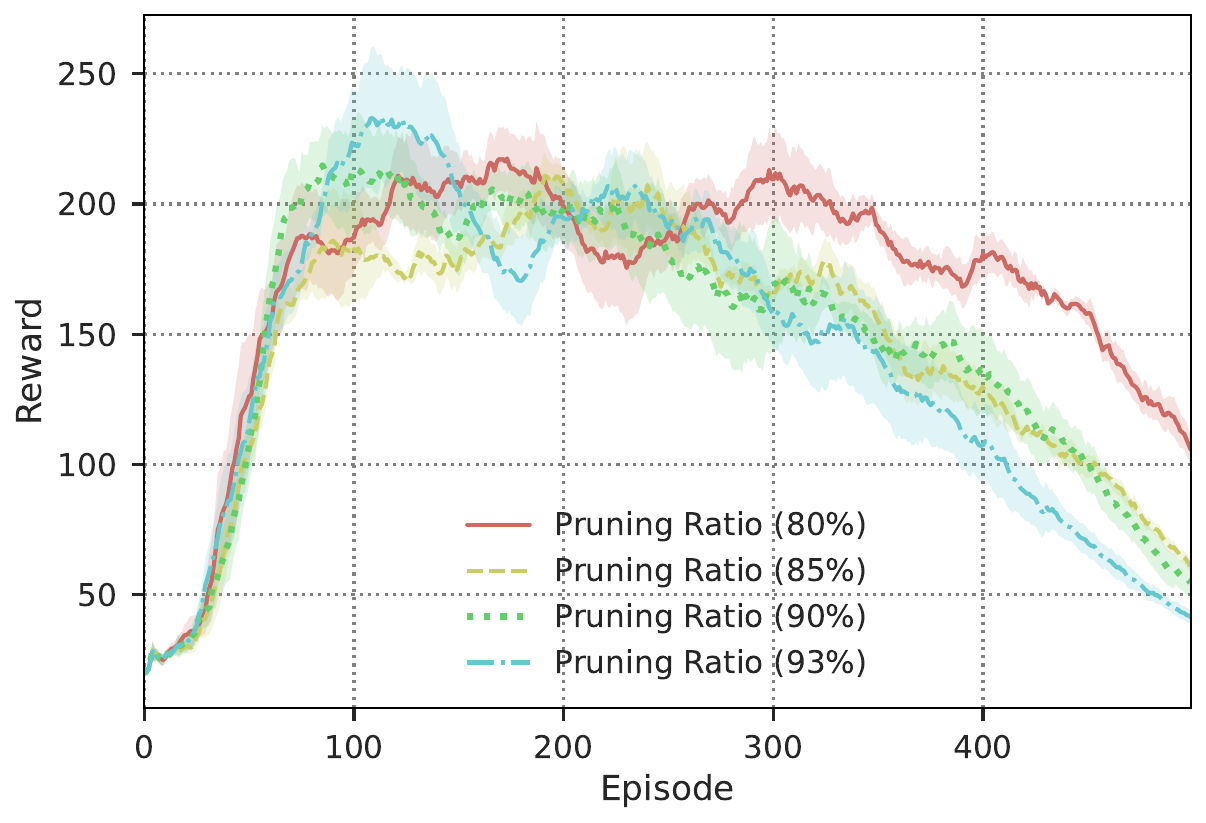}}
\subfloat[]{
\includegraphics[width=1.8in]{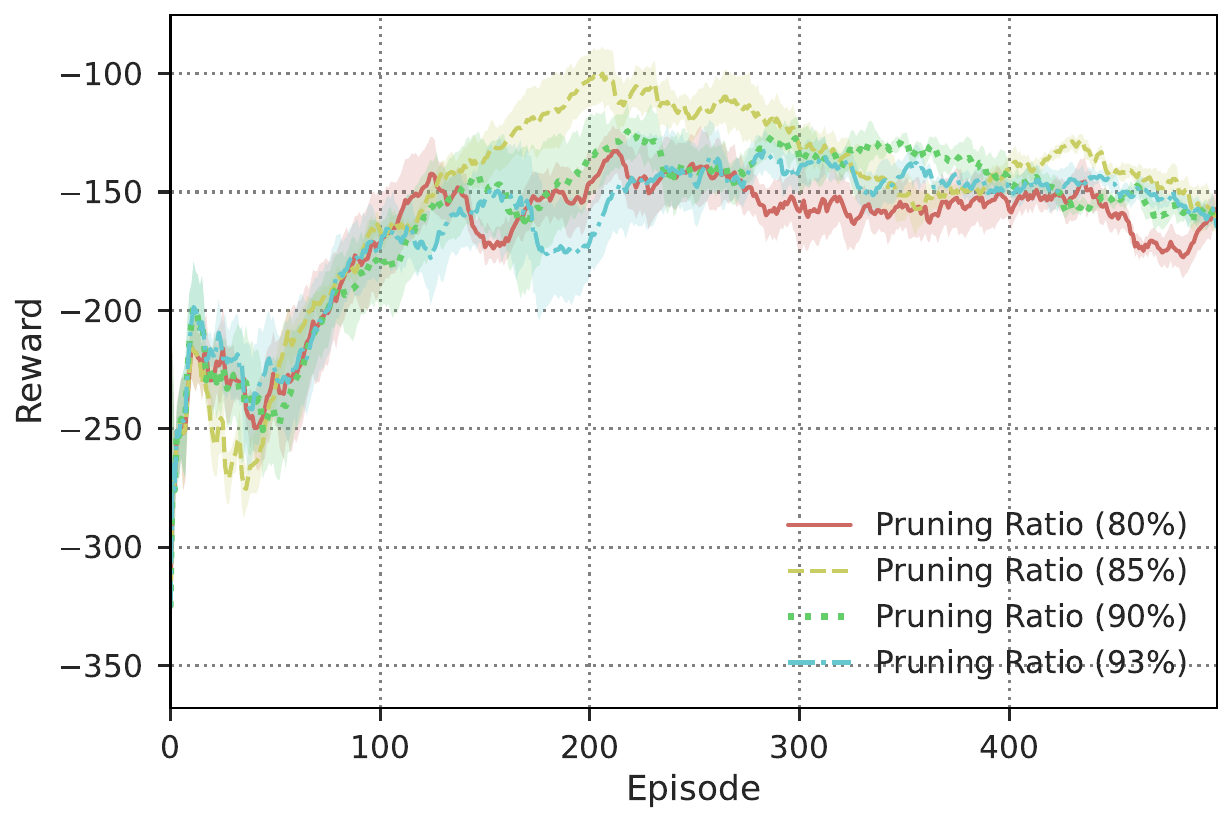}}
\subfloat[]{
\includegraphics[width=1.8in]{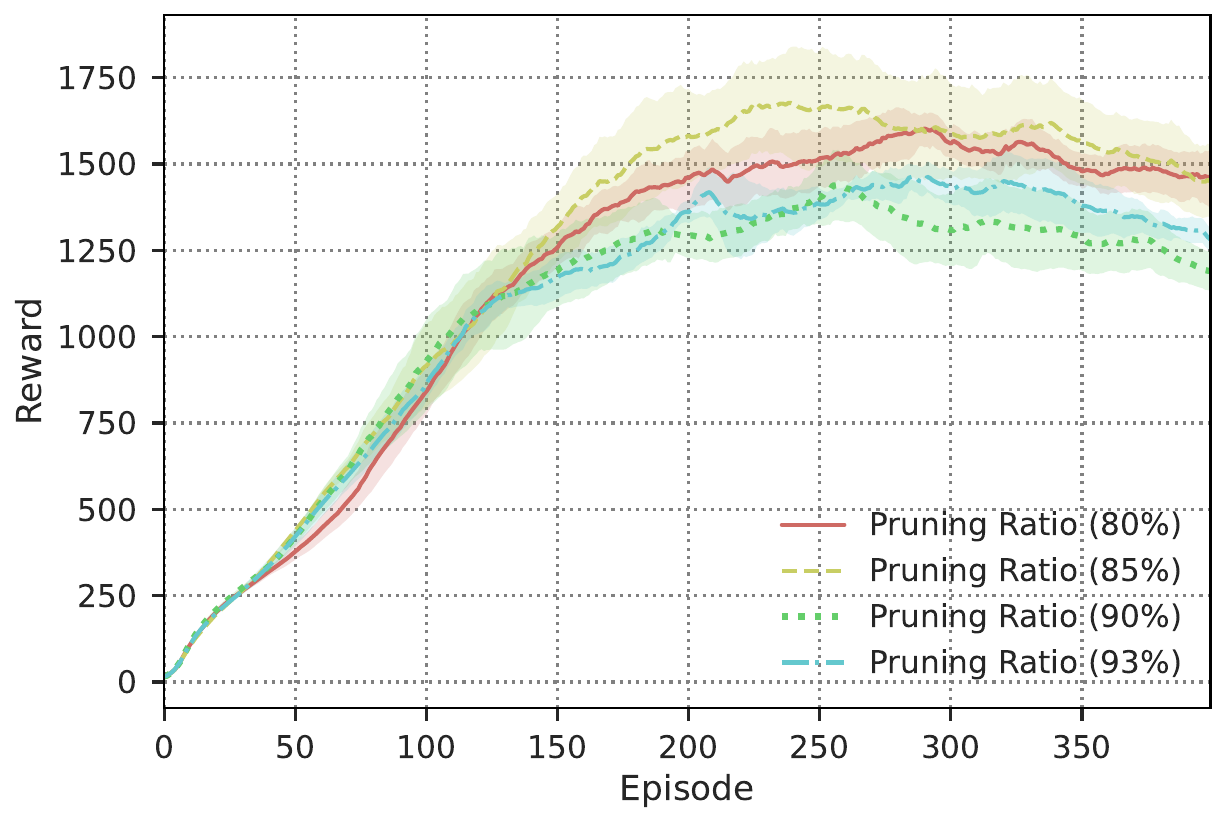}}
\subfloat[]{
\includegraphics[width=1.8in]{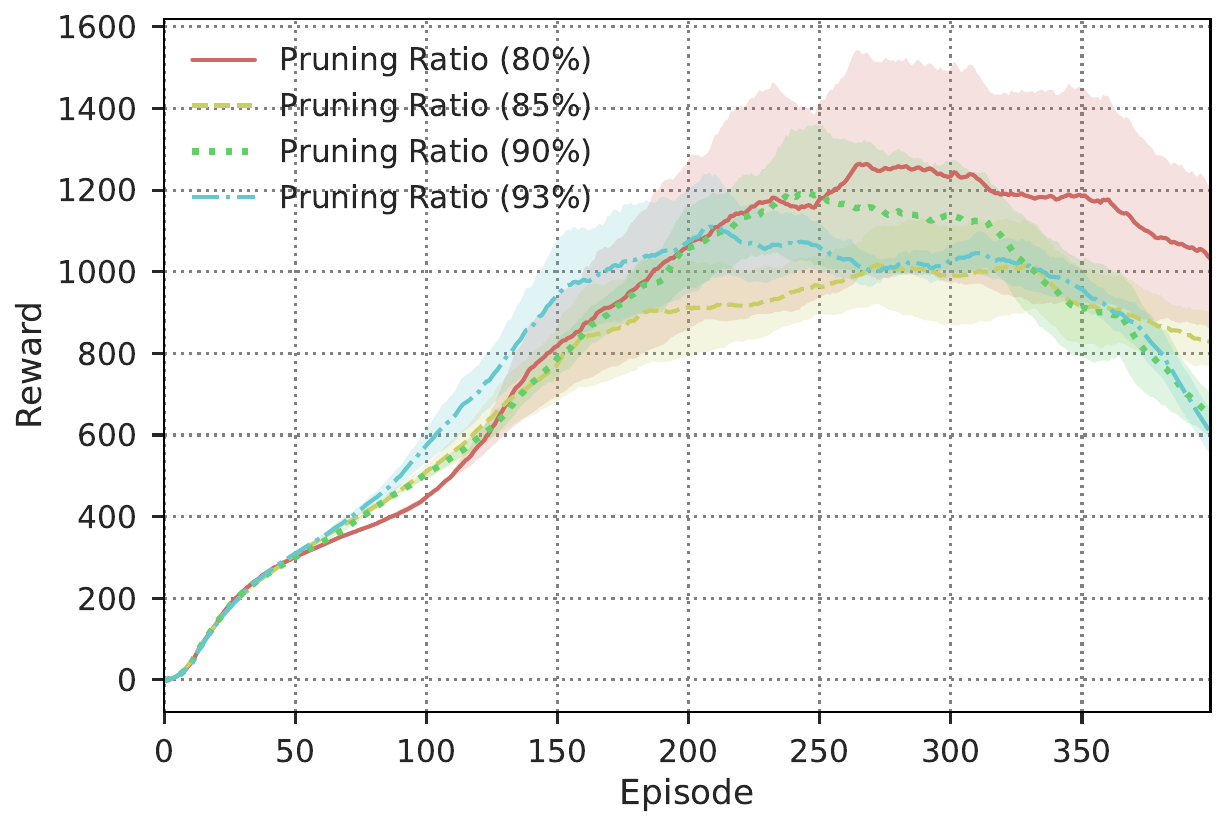}}
\end{center}
\caption{Performances of the PPO-Dropout method with a range of dropout ratio $r$ in DRL environments. (a) Cart Pole. (b) Lunar Lander. (c) Hopper. (d) Walker2D.}
\label{fig:E1-random}
\end{figure*}

\begin{figure*}[!t]
\begin{center}
\subfloat[]{
\includegraphics[width=1.8in]{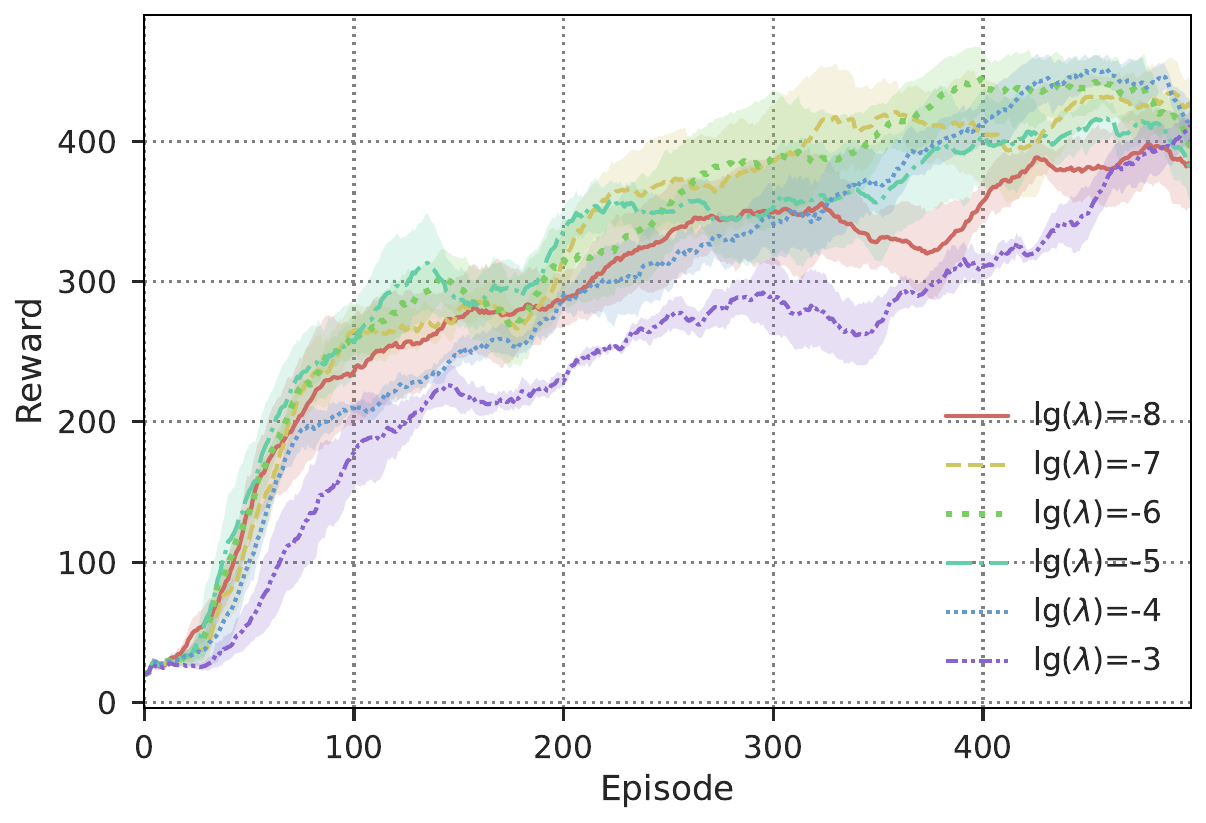}
}
\subfloat[]{
\includegraphics[width=1.8in]{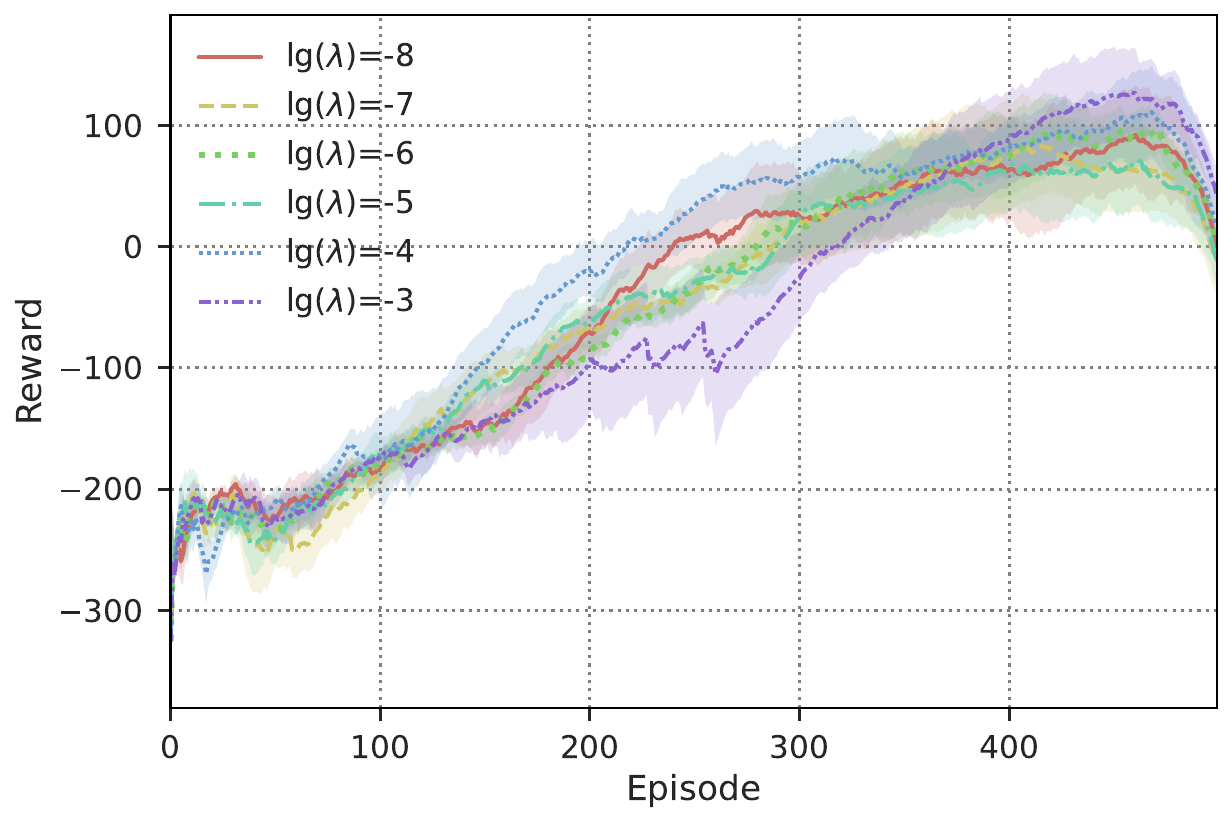}}
\subfloat[]{
\includegraphics[width=1.8in]{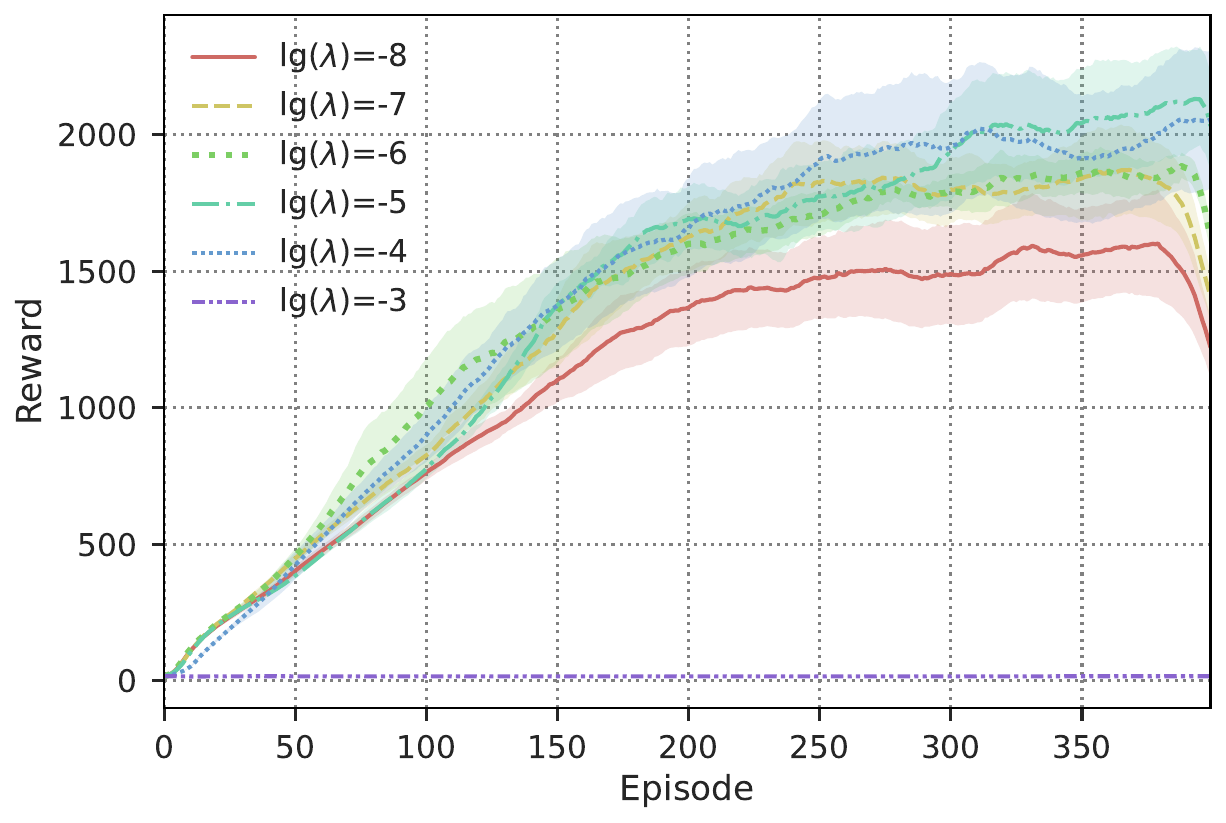}}
\subfloat[]{
\includegraphics[width=1.8in]{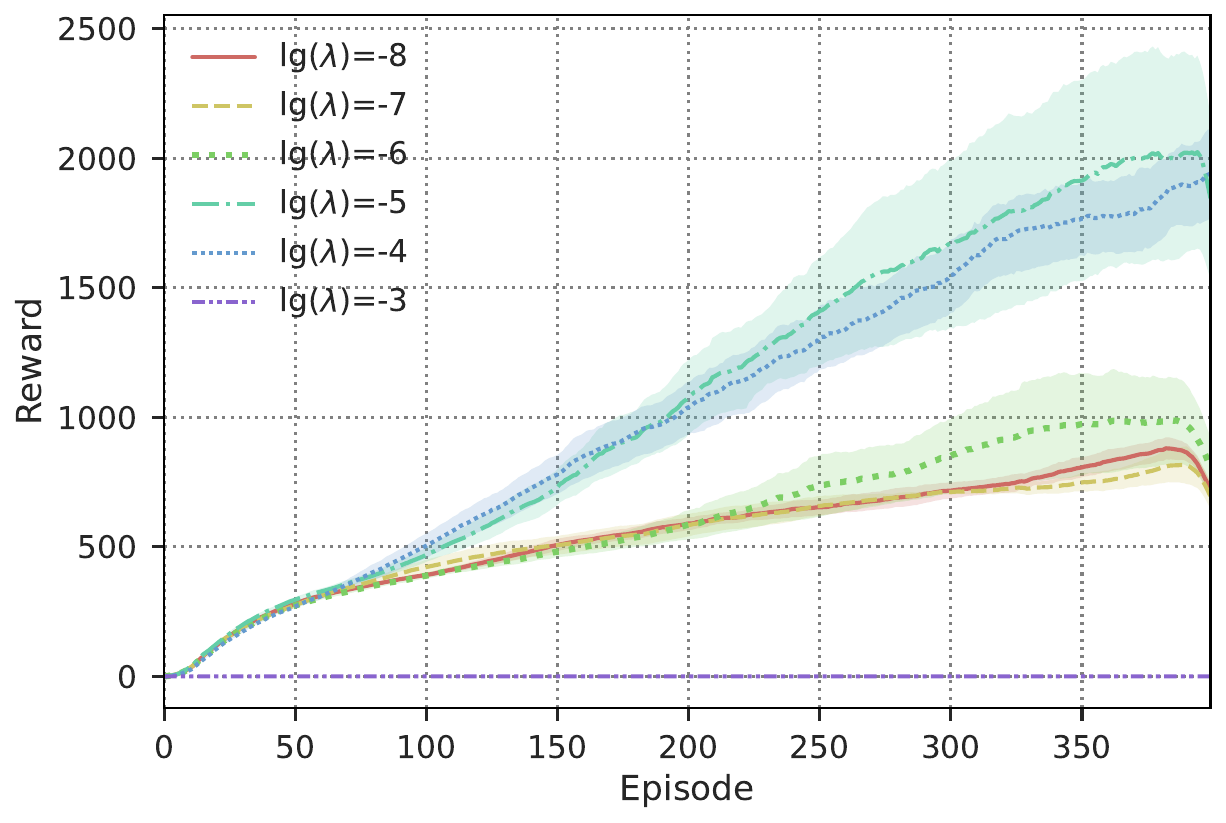}}
\end{center}
\caption{Performance results of the PPO-$l_1$ Lasso method in Cart Pole, Lunar Lander, Hopper, and Walker environments with different settings of $\lambda$. (a) Cart Pole. (b) Lunar Lander.
(c) Hopper. (d) Walker2D.}
\label{fig:E1-l1pro}
\end{figure*}

\begin{figure*}[!t]
\begin{center}
\subfloat[]{
\includegraphics[width=1.8in]{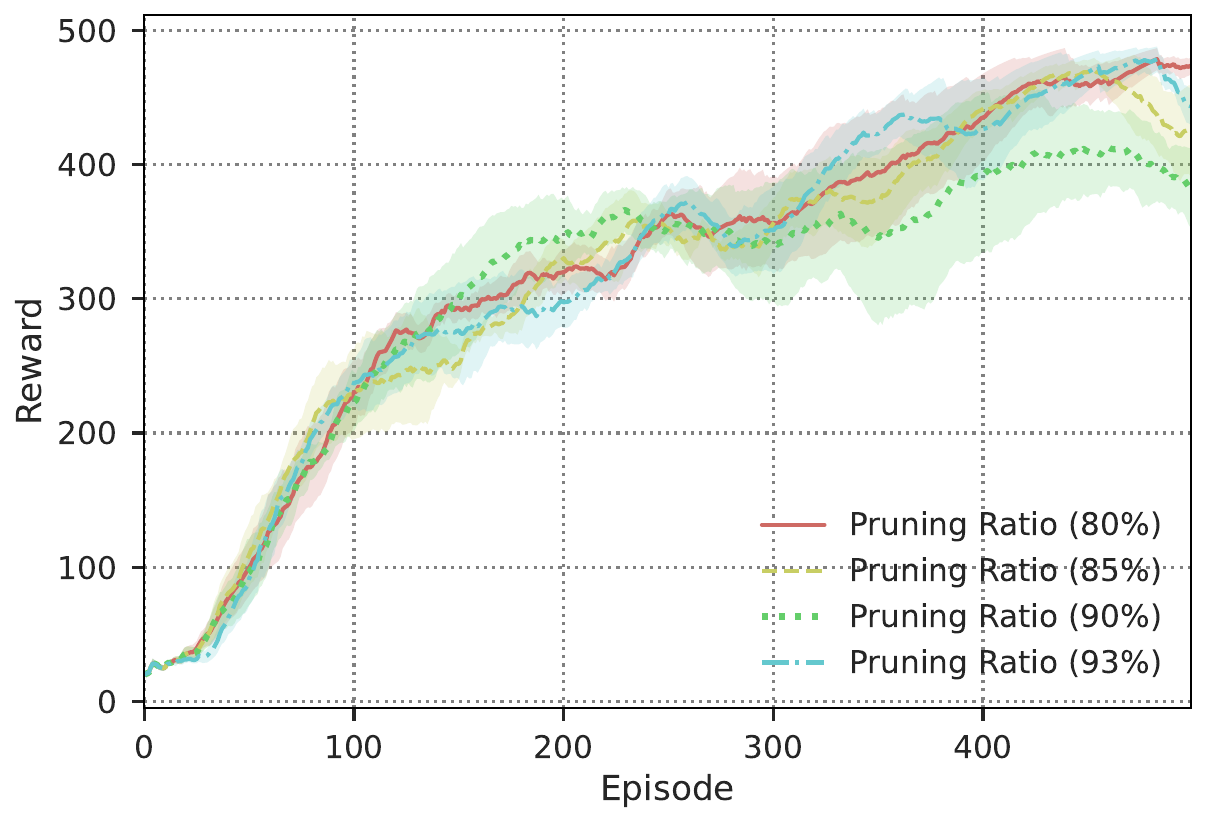}
}
\subfloat[]{
\includegraphics[width=1.8in]{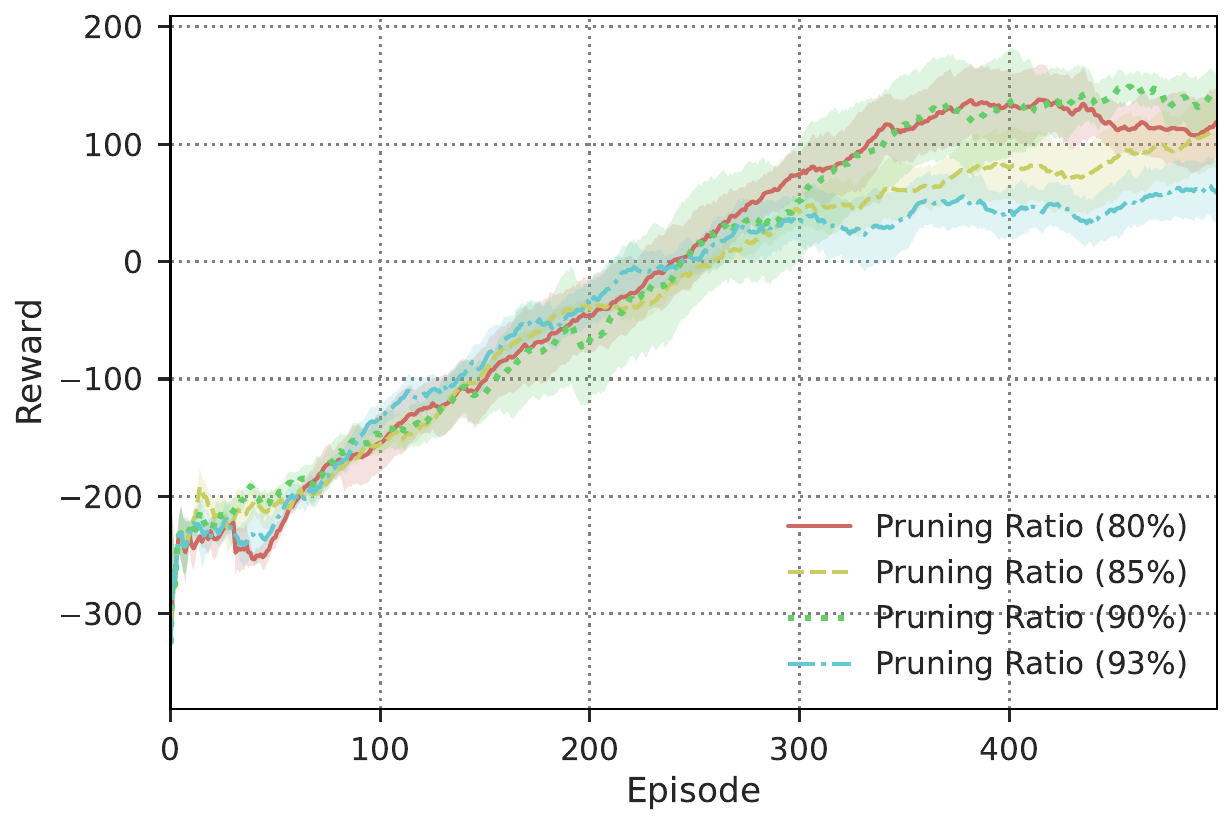}}
\subfloat[]{
\includegraphics[width=1.8in]{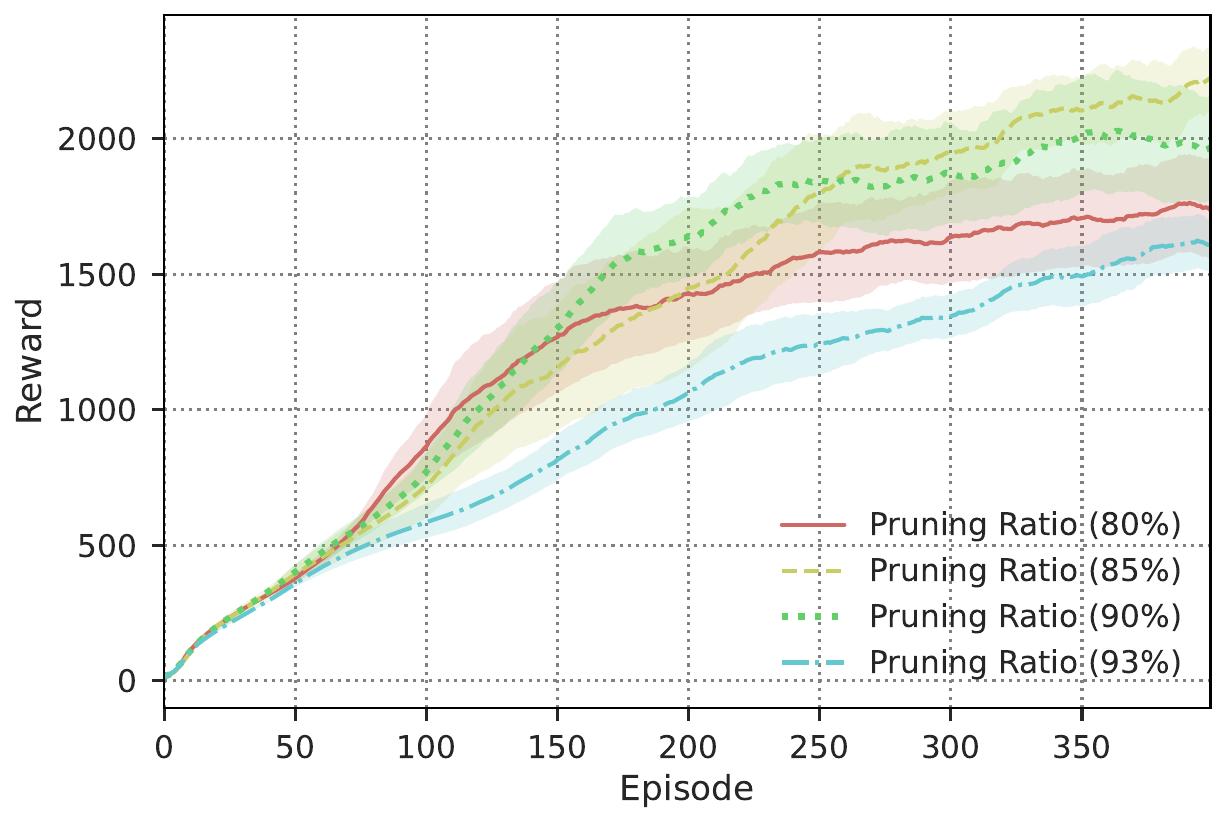}}
\subfloat[]{
\includegraphics[width=1.8in]{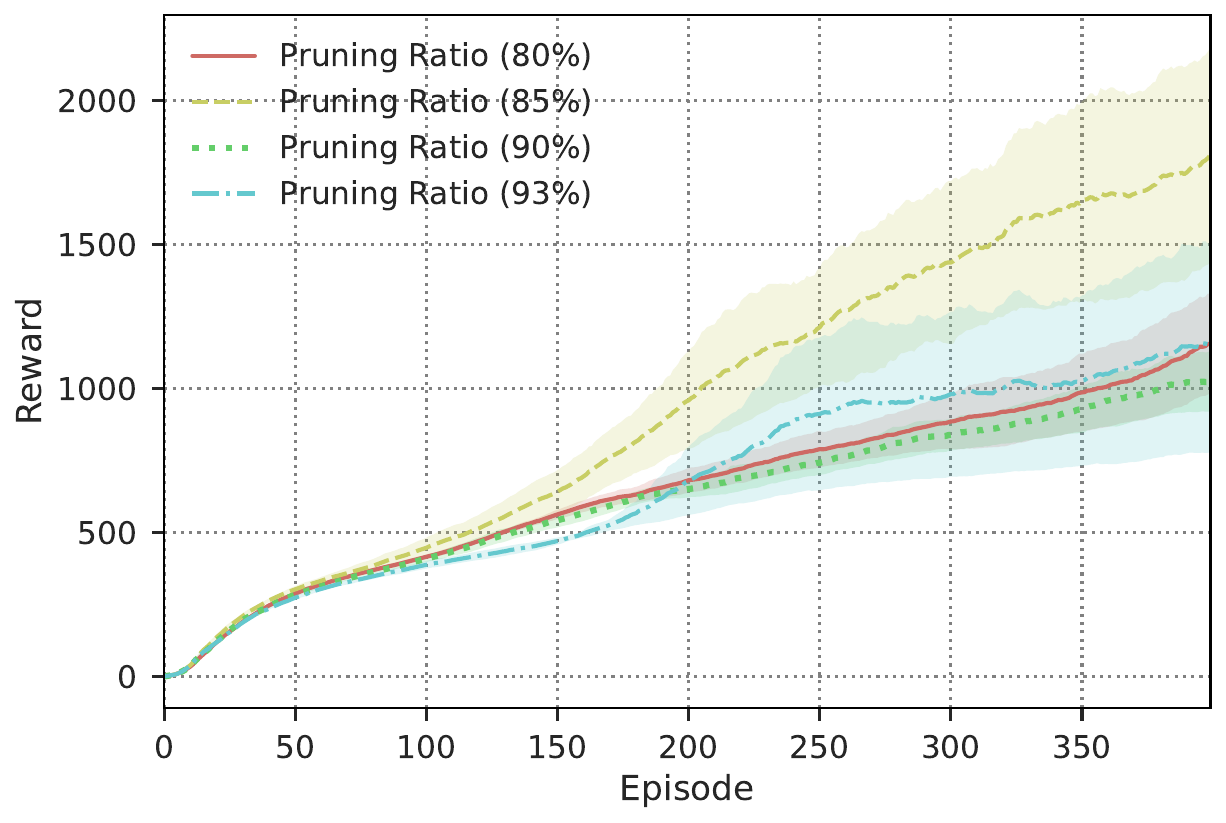}}
\end{center}
\caption{Performances of the PPO-PoPS method with a range of dropout ratio $r$ in DRL environments. (a) Cart Pole. (b) Lunar Lander. (c) Hopper. (d) Walker2D.}
\label{fig:E1-pops}
\end{figure*}

In this section, we present the experimental evaluation with parameters for PPO-SSL~\cite{wen2016learning}, PPO-Dropout~\cite{srivastava2014dropout}, PPO-$l_1$ Lasso~\cite{li2022compact}, PPO-PoPS~\cite{livne2020pops} and PPO-DSP, in terms of their performance under different sets of parameters.  All algorithms are run five times in four environments for all parameters to obtain comparable average performances with derivations.

The actor and critic networks both incorporate two fully connected hidden layers, and each hidden layer contains 128 neurons. In the Cart Pole environment, the input layer of the actor network is composed of four neurons, and the output layer contains one neuron. In the Lunar Laner environment, the input layer of the actor network has eight neurons, and the output layer consists of four neurons. In the Hopper environment, the input layer consists of 11 neurons, corresponding to the state observations, and the output neurons are four. Similarly, in the Walker2D environment, the environment is more complex because the agent controls a two-legged figure. The input layer consists of 17 neurons, and the output neurons are six.

\cref{fig:E1-con} shows hyperparameters tuning results of the PPO-DSP method in four DRL environments. Since PPO-DSP has two hyperparameters to adjust, they are $\mu$ and the pruning ratio, where $\mu$ is the penalty factor and the pruning ratio is the percentage of pruned parameters to the total parameters of the unpruned model. Finding the optimal pruning ratio and $\mu$ to achieve the balance between sparsity and accuracy is a challenging task, the adjustment range of $\mu$ is set to $\{10^{-8},10^{-7},10^{-6},10^{-5},10^{-4},10^{-3}\}$ and the pruning ratio is $\{80\%,85\%,90\%,93\%\}$. As can be seen in~\cref{fig:E1-con}(a), With the increase of pruning ratio, the reward has a tendency to decrease. When $\mu$ is $-7$ and the pruning ratio is $93\%$, the reward is approximately $272.99 \pm 124.96$, the PPO-DSP method has a good trade-off between performance (reward) and compression ratio (93\%) in the CartPole environment. \cref{fig:E1-con}(b) shows that when $\mu$ is $-8$ and the pruning ratio is 80\%, the PPO-DSP method has the maximum reward $-19.31 \pm 197.22$. When the pruning ratio is $93\%$, the PPO-DSP method achieves the maximum reward ($-28.09 \pm 157.55$), and the corresponding $\mu$ is $-7$ in the LunarLander environment. As shown in~\cref{fig:E1-con}(c)-(d), at the highest pruning rate ($93\%$), the PPO-DSP method achieves maximum rewards $1476.62 \pm 997.43$ and $1310.67 \pm 899.28$ in the Hopper and Walker environments.

As shown in \cref{fig:E1-glasso}, the PPO-SSL method also has two hyperparameters that need to be adjusted, i.e., $\mu$ and pruning ratio, and the experimental setting is consistent with the PPO-DSP method. According to the experimental results in~\cref{fig:E1-glasso}(a), the overall performance of the PPO-SSL method is close to that of the PPO-DSP method, and the average reward ($239.08 \pm 127.72$) is significantly lower than that of the PPO-DSP method when the pruning rate increases to $93\%$. Similarly,~\cref{fig:E1-glasso}(b)-(d) shows that the PPO-SSL method achieves the maximum reward $-44.22 \pm 119.13$, $1158.74 \pm 844.18$, $1098.34 \pm 842.67$ in the LunarLander, Hopper, and Walker environments respectively. The reward of the proposed PPO-DSP method is higher than that of the PPO-SSL method, especially when the pruning ratio is relatively high. This suggests that although the PPO-SSL method considers the individual relationships of neurons, it ignores the inter-layer relationships of neurons, which leads to the possibility that important neurons may be mispruned during the pruning process.

PPO-SGS applies group $l_1$ sparsity constraints on the first fully connected layer of the deep neural network to select the most important neurons, which reduces input dimensionality while aiming to improve classification accuracy.~\cref{fig:E1-l21norm} shows performance results of the PPO-SGS method on various DRL environments. As can be seen in~\cref{fig:E1-l21norm}(a), it shows that PPO-SGS obtained a mean cumulative reward of 268.26. This underperformed uncompressed PPO at 449.22 and PPO-PoPS at 421.40, but exceeded PPO-SSL at 239.08, PPO-Dropout at -52.22 and PPO-L1 Lasso at 402.82. In the more complex LunarLander environment,~\cref{fig:E1-l21norm}(b) shows that PPO-SSL achieved the highest mean cumulative reward at -128.09 among all hyperparameters. PPO-SGS appears to learn policies inefficiently in this environment. Its final performance surpasses even the uncompressed PPO model (-61.03). This indicates that PPO-SGS's aggressive pruning technique may provide an implicit regularization effect to prevent overfitting. For the continuous control tasks Hopper and Walker2D,~\cref{fig:E1-l21norm}(c)-(d) shows that PPO-SGS substantially exceeds the performance of all baseline methods. Over the last 30 episodes, it attains average rewards of 1476.62 and 1310.67 respectively. Additionally, it exceeds the baseline rewards of 1158.74 (PPO-SSL), 1445.25 (PPO-Dropout), 1396.63 (PPO-$l_1$ Lasso) for Hopper, and 1098.34 (PPO-SSL), 1002.57 (PPO-Dropout), 2032.92 for Walker2D.

Different from the previous three structure pruning methods, PPO-Dropout uses a random pruning strategy, which reduces the number of model parameters by randomly pruning unimportant neurons during the training process. Since PPO-Dropout does not impose penalty constraints on the weights, we set $\mu$ to zero and only adjust the hyperparameter pruning ratio.~\cref{fig:E1-random}(a) shows that when the PPO-Dropout method reaches 400 episodes, the reward decreases significantly, which suggests that ignoring the importance of neurons and randomly discarding neurons may lead to DRL model performance loss, and it is difficult to recover the performance.~\cref{fig:E1-random} (b)-(d) shows that the reward of the PPO-Dropout method is significantly worse than that of the PPO-DSP method when the pruning ratio is $93\%$, and the reward is $-144.36 \pm 9.47$, $1445.25 \pm 155.49$, and $1002.57 \pm 201.99$ respectively in Lunarlander, Walker, and Hopper environments.

PPO-$l_1$ Lasso and PPO-PoPS are unstructured pruning methods. The object of their pruning is weight, which is realized by adding masks to unimportant weights. Although the unstructured pruning method can not prune the whole neuron, it can make the weights connected to any side of the neuron all be zero, so as to achieve the effect of structured pruning. Since PPO-$l_1$ Lasso is an adaptive pruning technique, the hyperparameter that needs to be adjusted is $\mu$. As shown in~\cref{fig:E1-l1pro}, PPO-$l_1$ Lasso achieves rewards in the CartPole, LunarLander, Walker, Hopper are $402.82 \pm 10.11$, $-49.43 \pm 47.11$, $2032.92 \pm 497.54$, $1396.63 \pm 597.68$, respectively, and the corresponding $\mu$ are $-4$, $-3$, $-5$, $-5$, respectively. When the model weights are constantly pruned during the training iterations, PPO-$l_1$ Lasso will have an obvious downward trend at the end of the training process. This is because the retained weights in the final training stage are of high importance, and the performance of the model will be affected if pruned. 

Since PPO-PoPS is the hard threshold pruning method, we adjust the hyperparameter of the pruning rate.~\cref{fig:E1-pops} shows that the convergence curve of the PPO-Dropout method is relatively smooth throughout the training process. PPO-PoPS achieves the best rewards in CartPole, LunarLander, Hopper, and Walker are $442.77 \pm 10$, $134.33 \pm 19.28$, $2198 \pm 97.36$, $1428.67 \pm 496.58$, respectively.

\subsection{Performance Comparison}
\begin{figure*}[!t]
\begin{center}
\subfloat[CartPole-v1.]{
\includegraphics[width=2.5in]{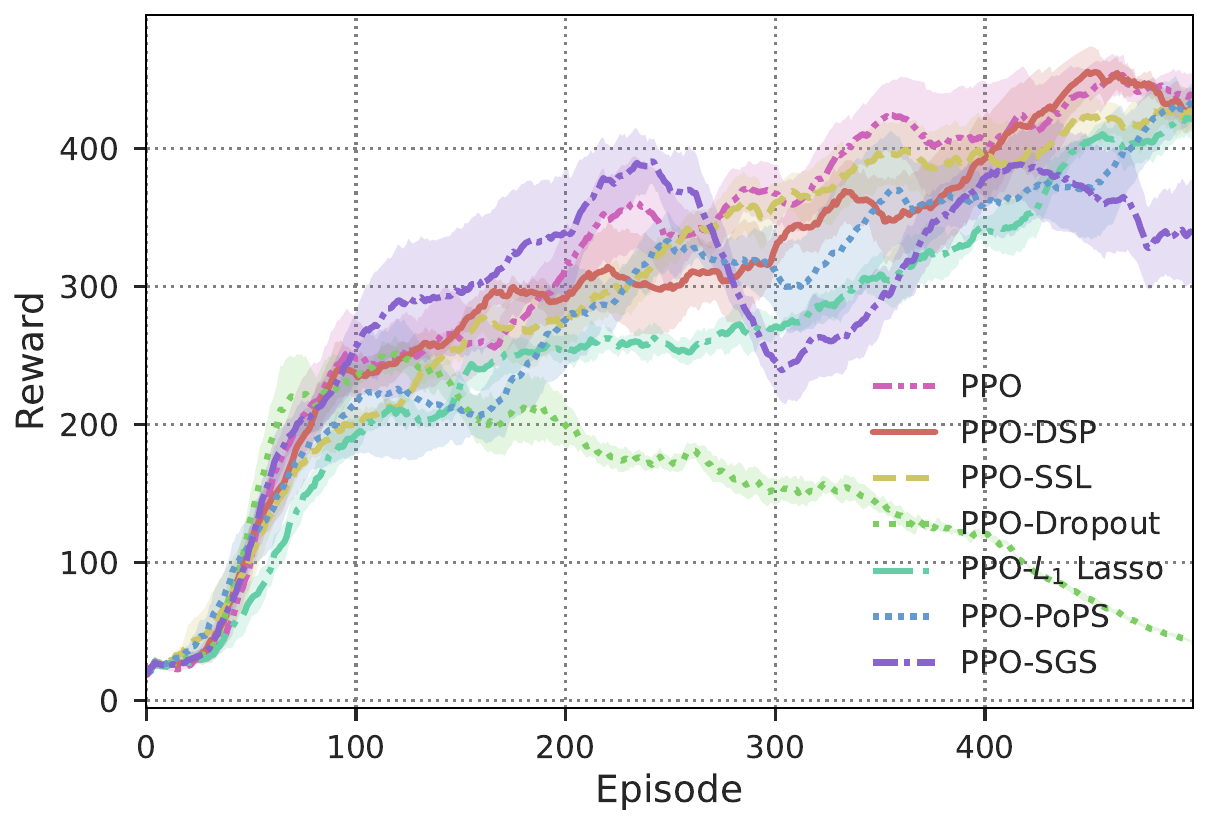}
}
\subfloat[LunarLander-v2.]{
\includegraphics[width=2.5in]{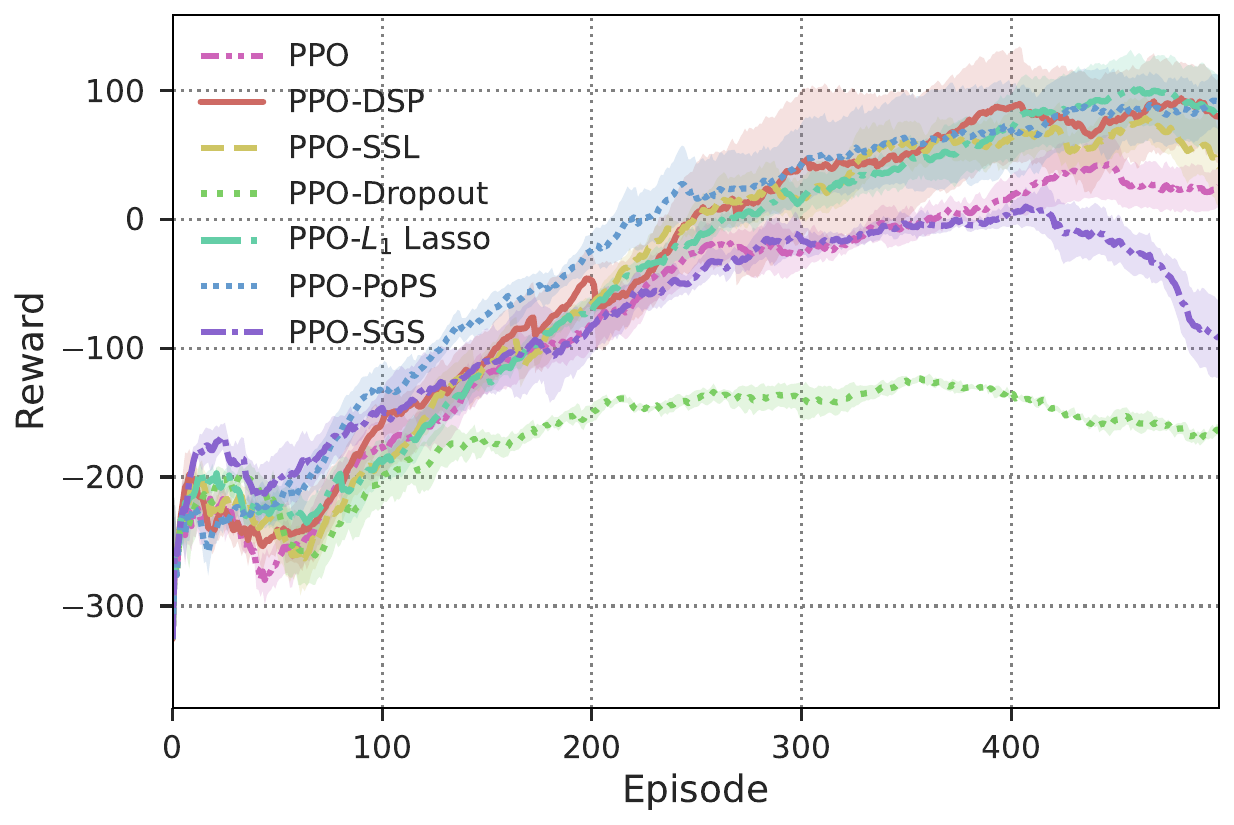}}

\subfloat[Hopper-v3.]{
\includegraphics[width=2.5in]{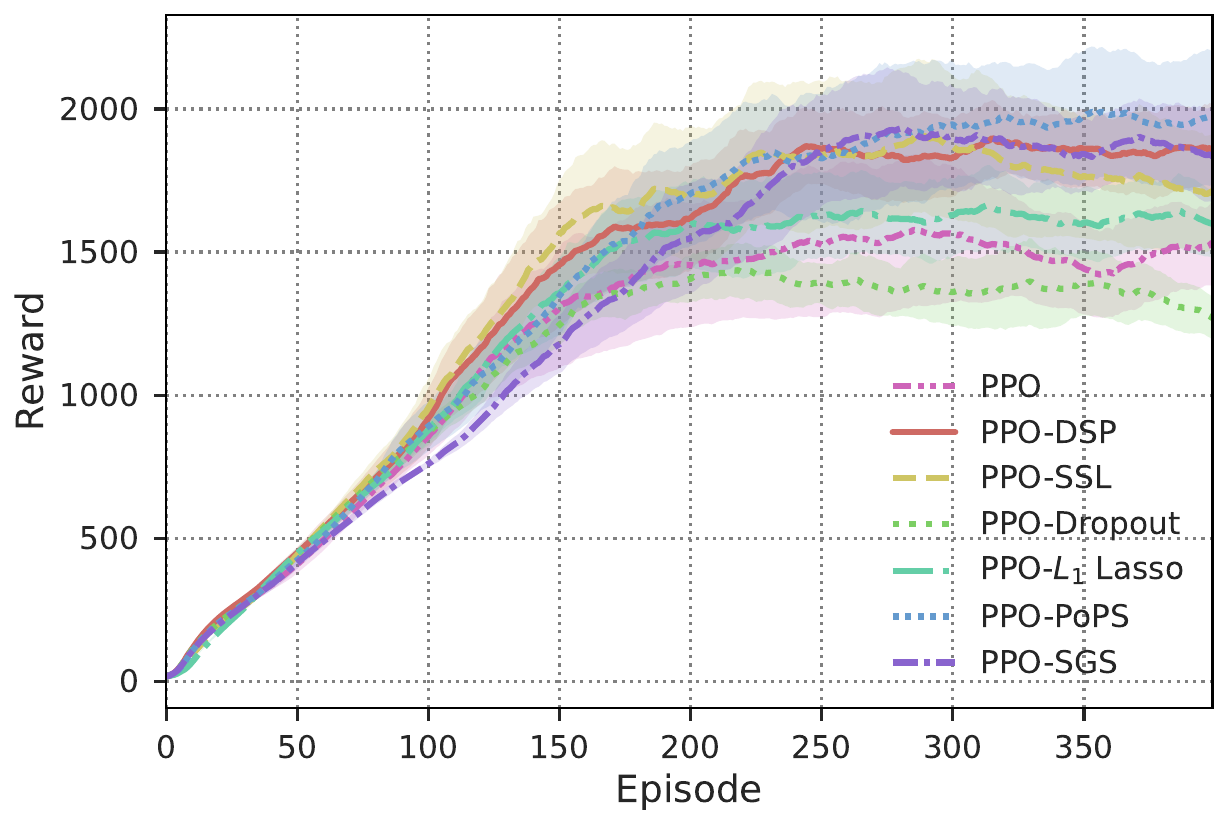}}
\subfloat[Walker2D-v3.]{
\includegraphics[width=2.5in]{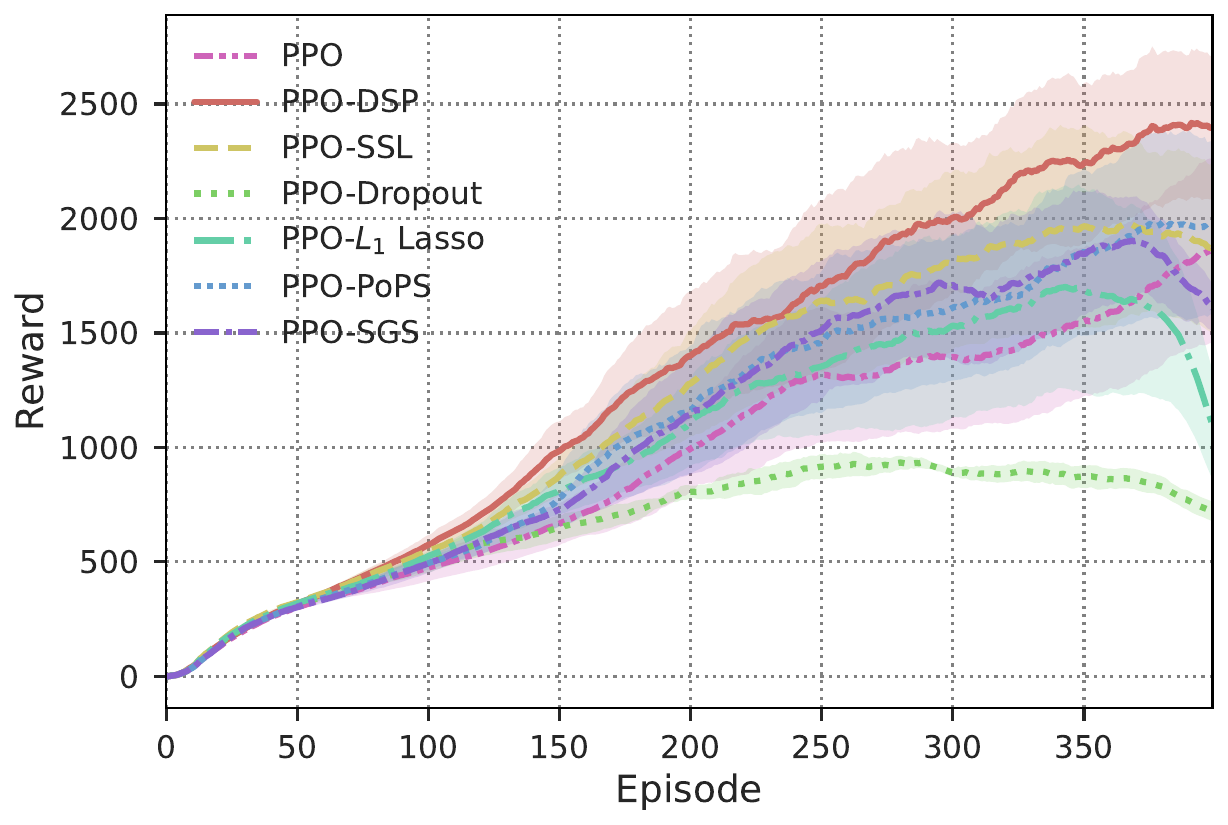}}
\end{center}
\caption{Learning curves of the baselines on different discrete and continuous DRL control tasks. The performance of each method is calculated as the average reward per episode over the last 30 evaluations of the training. The shaded region represents the standard deviation of the average evaluation over 5 runs.}
\label{fig:E2-total}
\end{figure*}

To demonstrate the effectiveness of the proposed algorithm, we compare the proposed algorithm with baselines under the best hyperparameter ($\mu$ or pruning ratio) and also compare it with the PPO algorithm. 


\cref{fig:E2-total} shows that the dynamic structured pruning method PPO-DSP demonstrates competitive performance compared to the baselines. In the CartPole environment,~\cref{fig:E2-total}(a) shows that PPO-DSP achieves a lower reward of 268.26 compared to uncompressed PPO at 449.22 and PPO-PoPS at 421.40. However, PPO-DSP outperforms PPO-SSL at 239.08, PPO-Dropout at -52.22, and PPO-L1 Lasso at 402.82. PPO-DSP achieves a lower reward than PPO and PPO-PoPS in the CartPole environment and also converges slower than other algorithms. One possible reason for this is that the CartPole environment is relatively simple and does not require a complex policy network to solve. Therefore, pruning the network may influence the Learning ability of the PPO-DSP model in a simple environment, leading to slower learning and lower performance. Another possible reason for the slower convergence rate is that the lighter network structure may make the policy more sensitive to the hyperparameters and the initialization of the network. This indicates PPO-DSP can maintain high performance while compressing the network with the $93\%$ pruning rate. In LunarLander environment,~\cref{fig:E2-total}(b) shows that PPO-DSP converges faster and achieves the highest reward at -28.09 among all methods. This demonstrates PPO-DSP's efficient policy learning ability in complex environments. The higher reward than PPO at -61.03 suggests PPO-DSP may have implicitly regularized the policy, preventing overfitting.
In the continuous control tasks Hopper and Walker2D,~\cref{fig:E2-total}(c)-(d) shows that PPO-DSP substantially outperforms the baselines, achieving rewards of 1476.62 and 2310.67. This compares to 1158.74 and 2098.34 for PPO-SSL, 1445.25 and 902.57 for PPO-Dropout, and 1396.63 and 1032.92 for PPO-L1 Lasso. Overall, PPO-DSP demonstrates strong performance across discrete and continuous tasks.

\subsection{Model Structure Visualization after Pruning}
\begin{figure*}[!t]
\begin{center}
\subfloat[The first layer weights in PPO (4 x 128)]{
\includegraphics[width=1.5in]{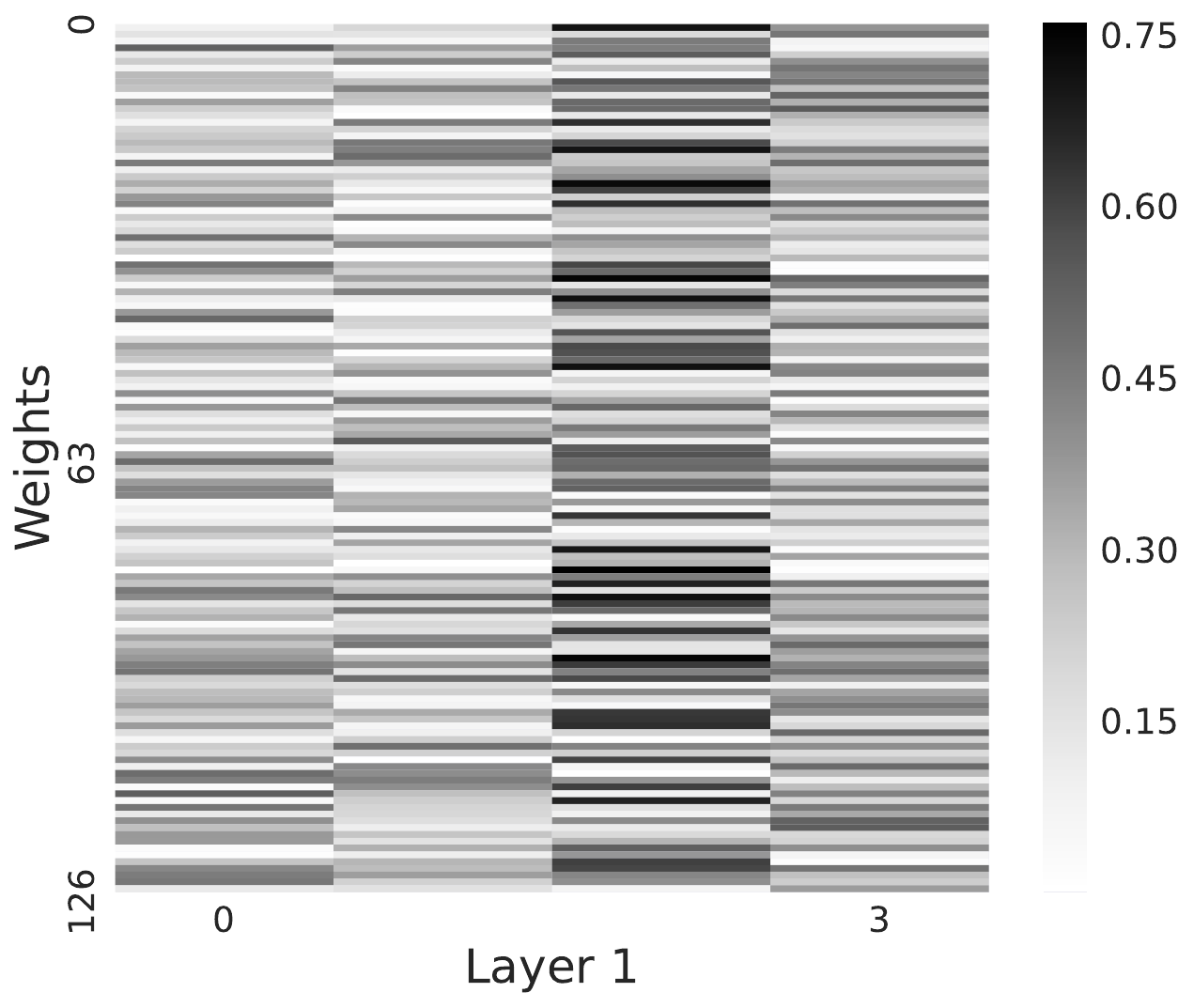}
}
\subfloat[The first layer weights in PPO-DSP (4 x 128)]{
\includegraphics[width=1.5in]{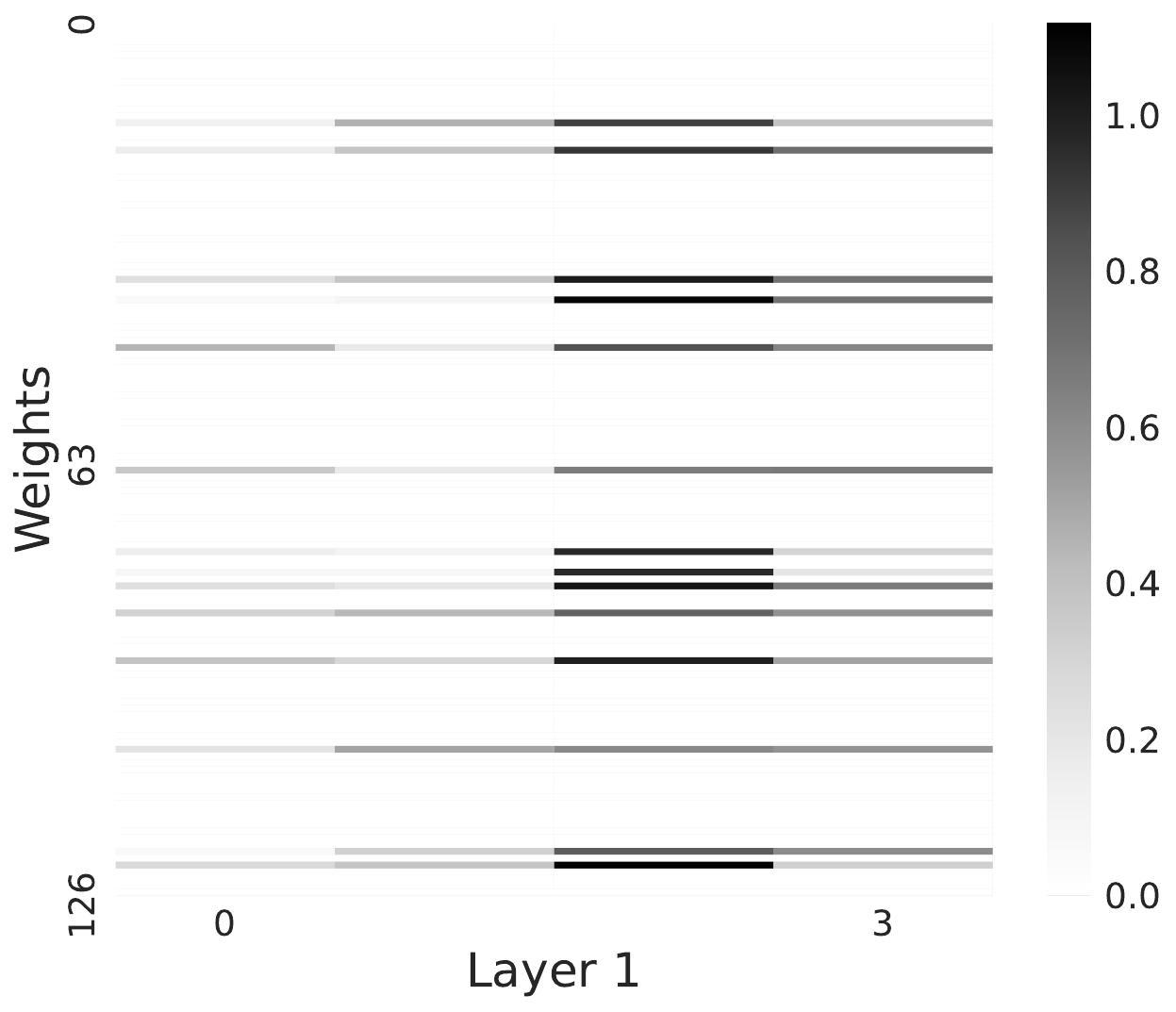}}
\subfloat[The first layer weights in PPO-SSL (4 x 128)]{
\includegraphics[width=1.5in]{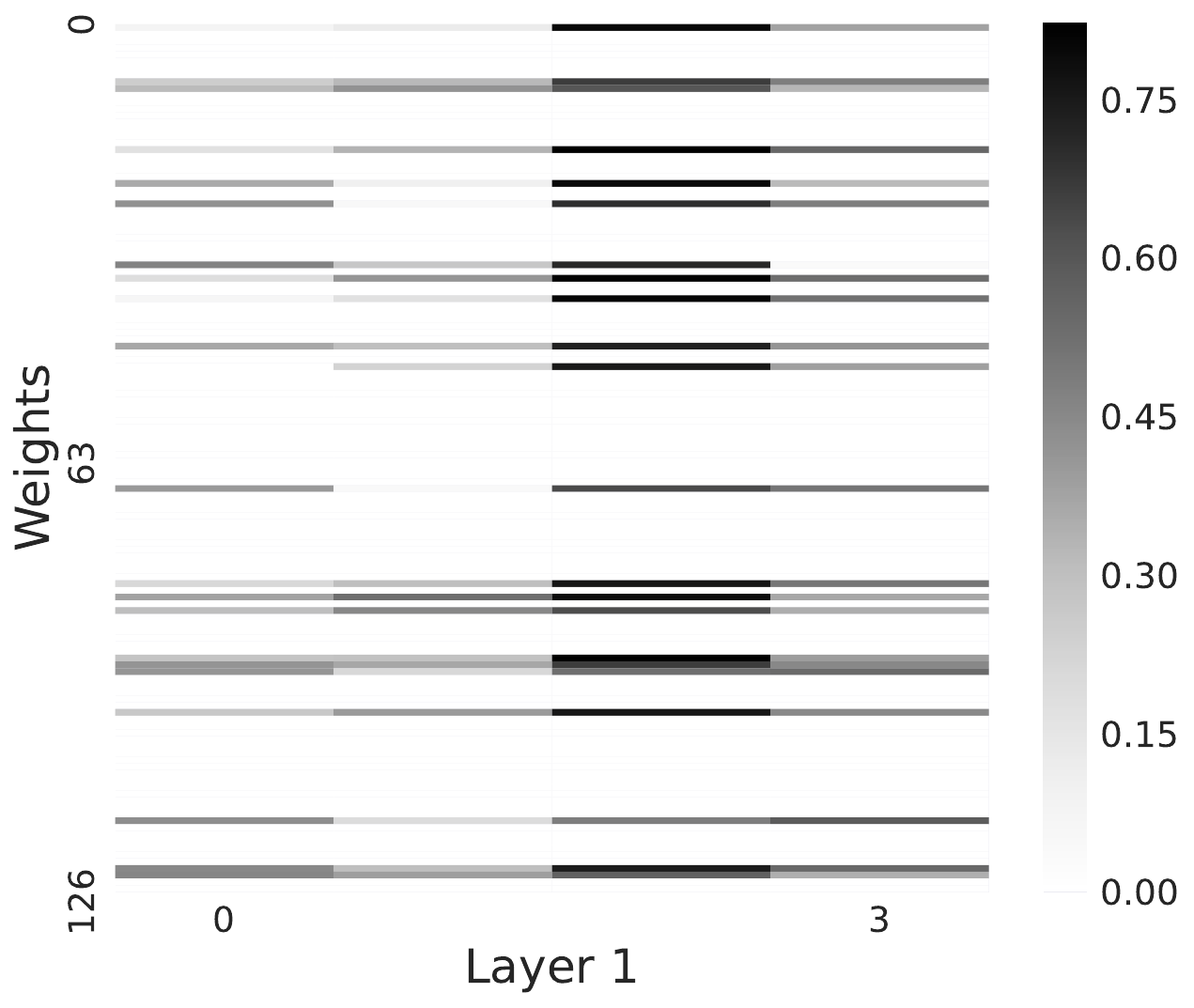}}
\subfloat[The first layer weights in PPO-$L_{1}$ Lasso (4 x 128)]{
\includegraphics[width=1.5in]{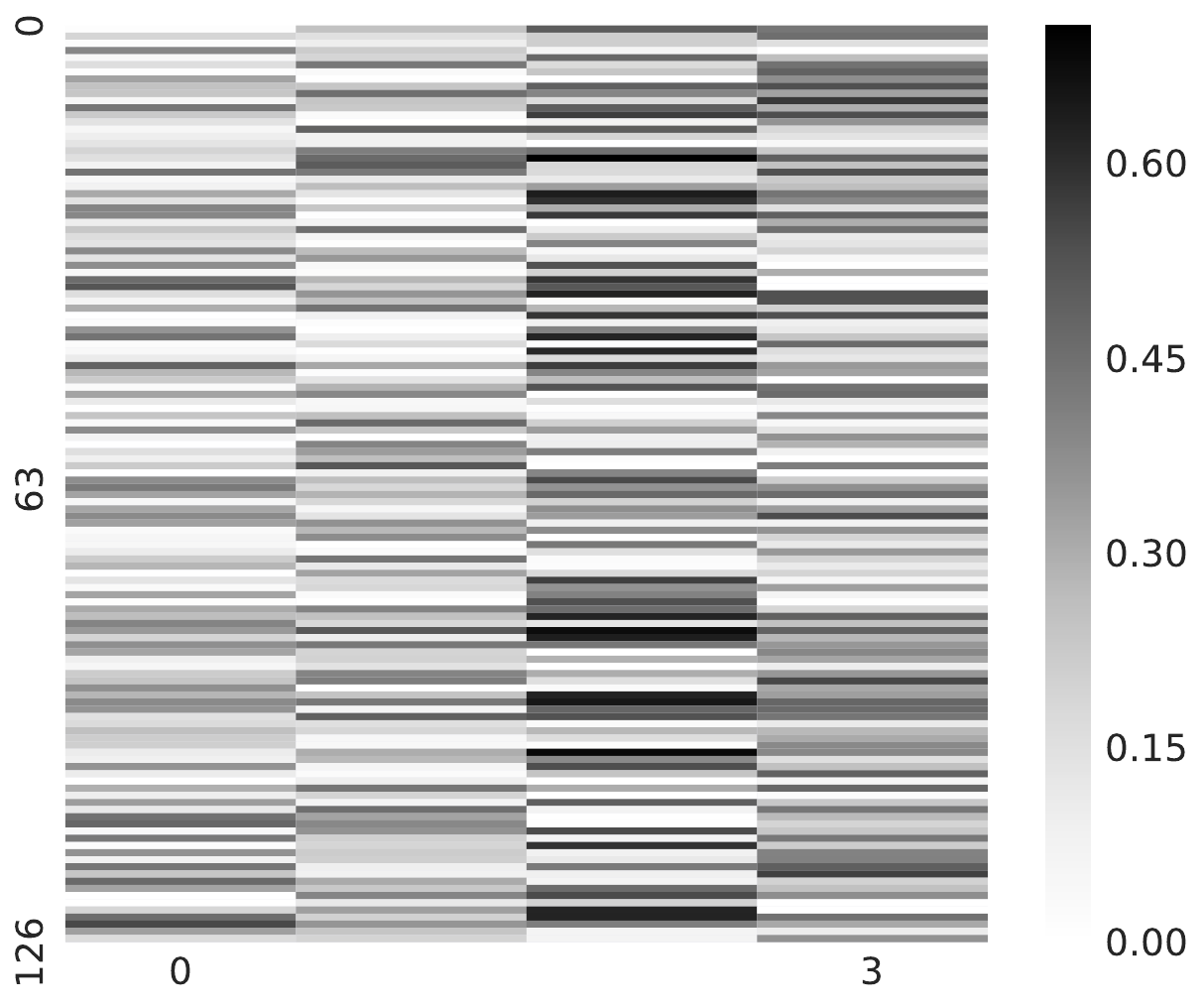}}\\
\subfloat[The second layer weights in PPO (128 x 128)]{
\includegraphics[width=1.5in]{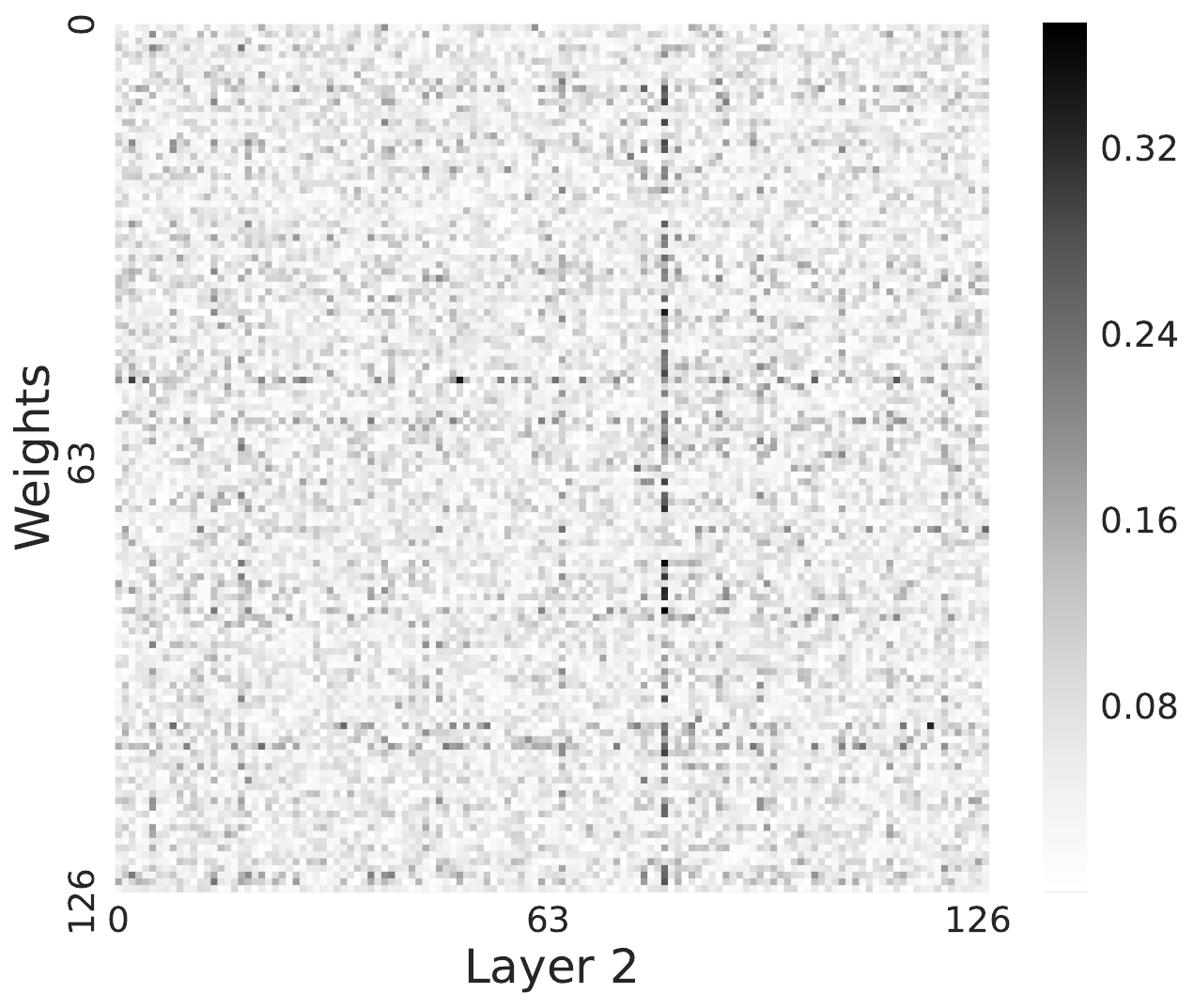}}
\subfloat[The second layer weights in PPO-DSP (128 x 128)]{
\includegraphics[width=1.5in]{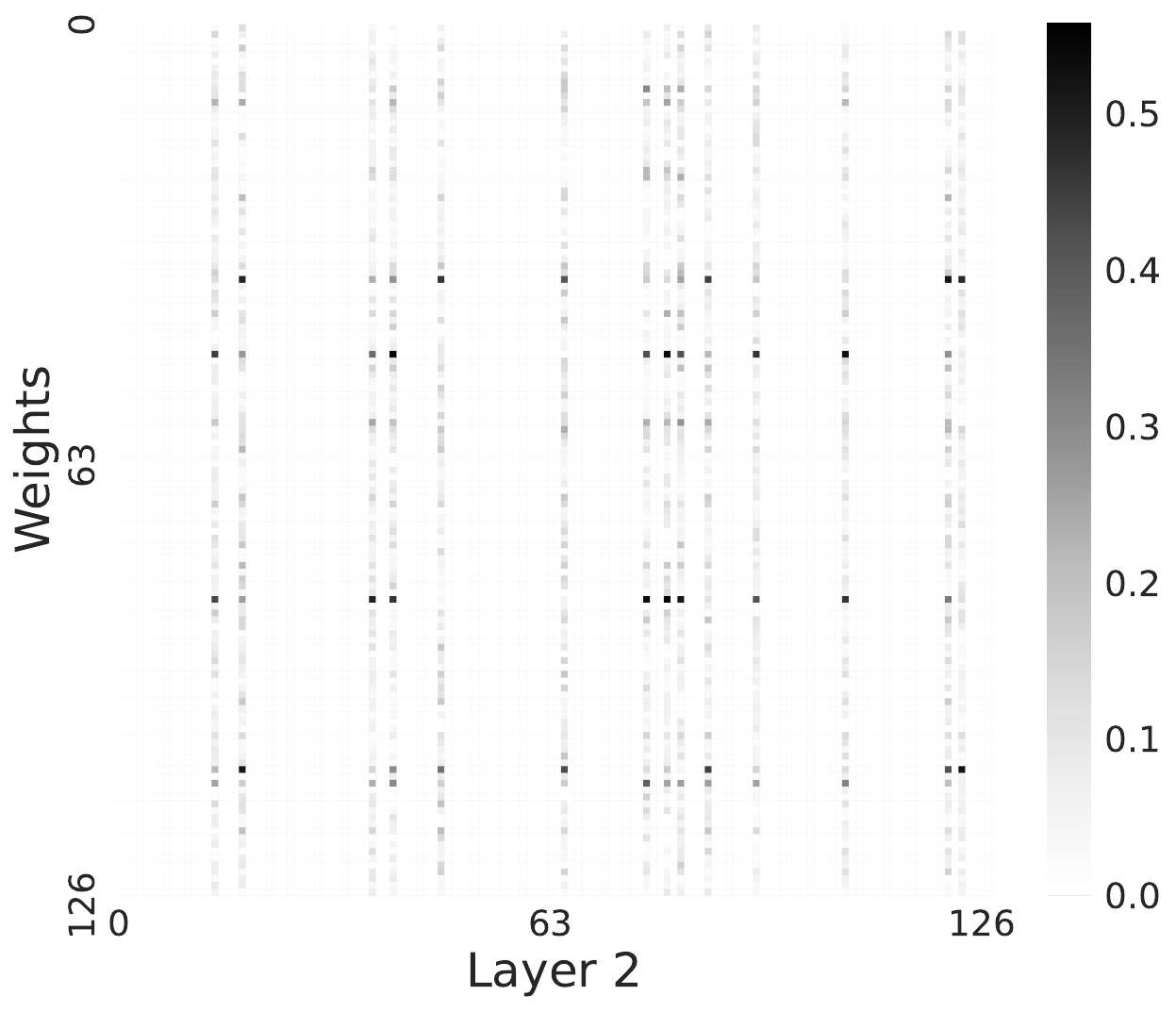}}
\subfloat[The second layer weights in PPO-SSL (128 x 128)]{
\includegraphics[width=1.5in]{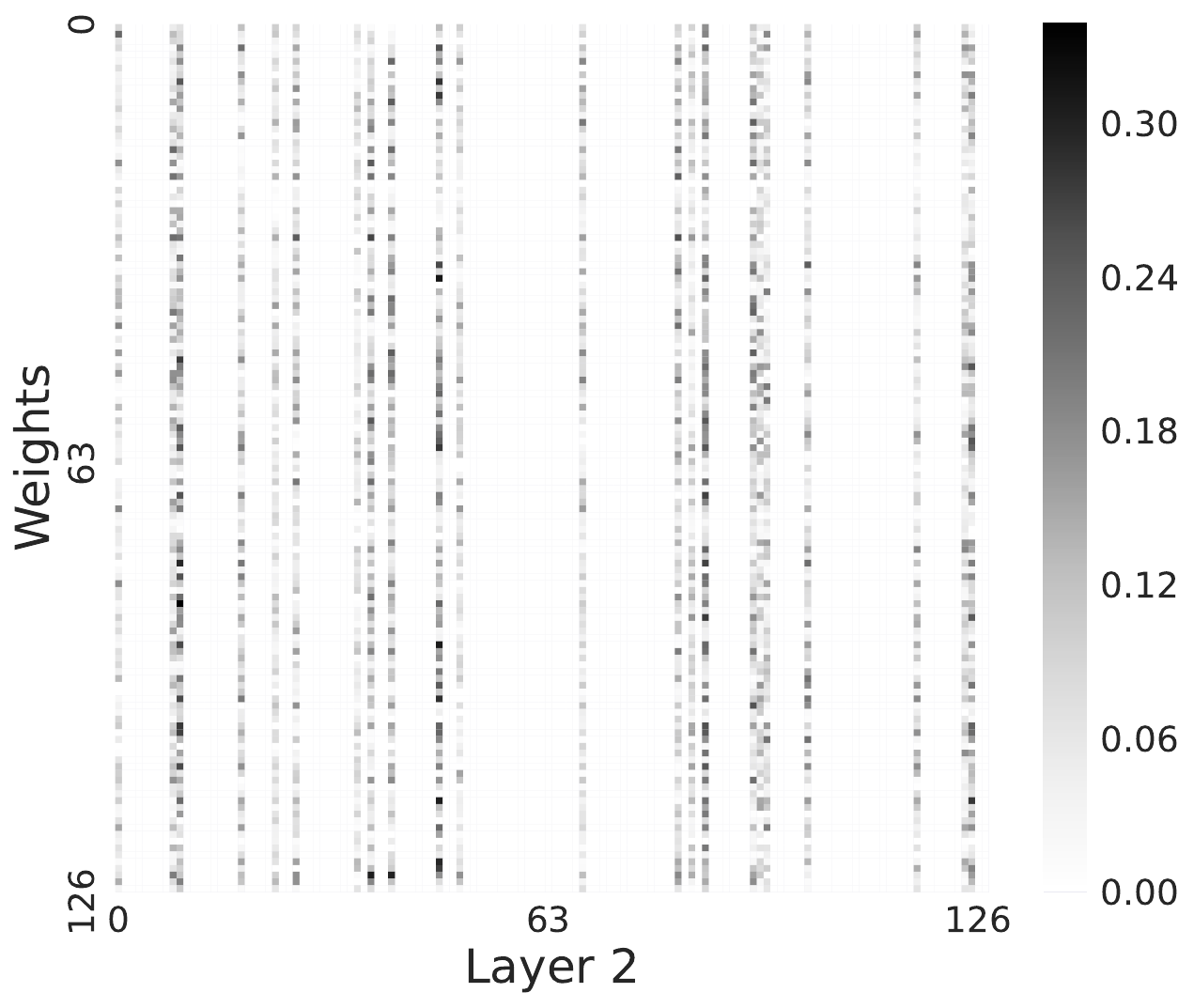}}
\subfloat[The second layer weights in PPO-$L_{1}$ Lasso (128 x 128)]{
\includegraphics[width=1.5in]{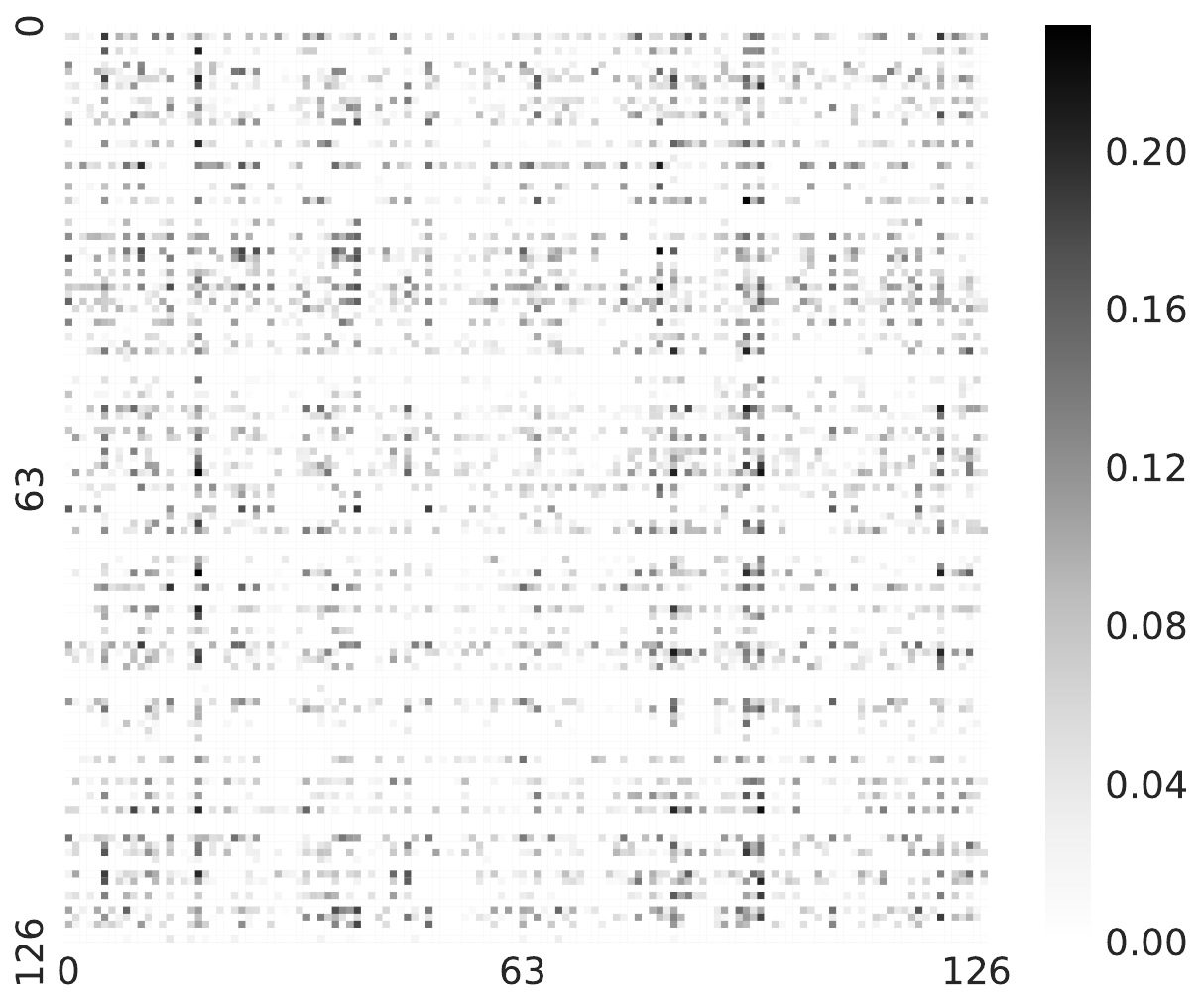}}
\end{center}
\caption{The absolute weight heat map visualization of baseline models. The row indicates the input layer of the model, while the column indicates the latter layer of the model.}
\label{fig:E3-heatmap}
\end{figure*}

To further analyze the effect of our proposed dynamic structured pruning method on the DRL model structure, we use heat map images to visualize the weight matrices of different layers of the actor network. In~\cref{fig:E3-heatmap}, the scale on the horizontal axis represents the sequence number of the neuron in the previous layer, and the scale on the ordinate represents the sequence number of the neuron in the current layer. Each gray square represents the weight value of the connection between the neurons in the previous layer and the neurons in the current layer. It can be seen that PPO-DSP can obtain more sparse weights while maintaining network performance, and structured sparse weights can be obtained by group sparse pruning, which means that some neurons are completely pruned and the corresponding columns or rows are all zero. On the contrary, PPO-SSL and PPO-$l_1$ Lasso still retain redundant sparse weight matrices, where many unimportant weights are closed to zero and not pruned. These weights increase the computational complexity and memory consumption of the network, and may also affect the network performance. Therefore, PPO-DSP can effectively reduce the network size and complexity by pruning the unimportant neurons and weights.

\subsection{Performance Evaluation}
\begin{figure*}[!t]
\begin{center}
\subfloat[]{
\includegraphics[width=2.5in]{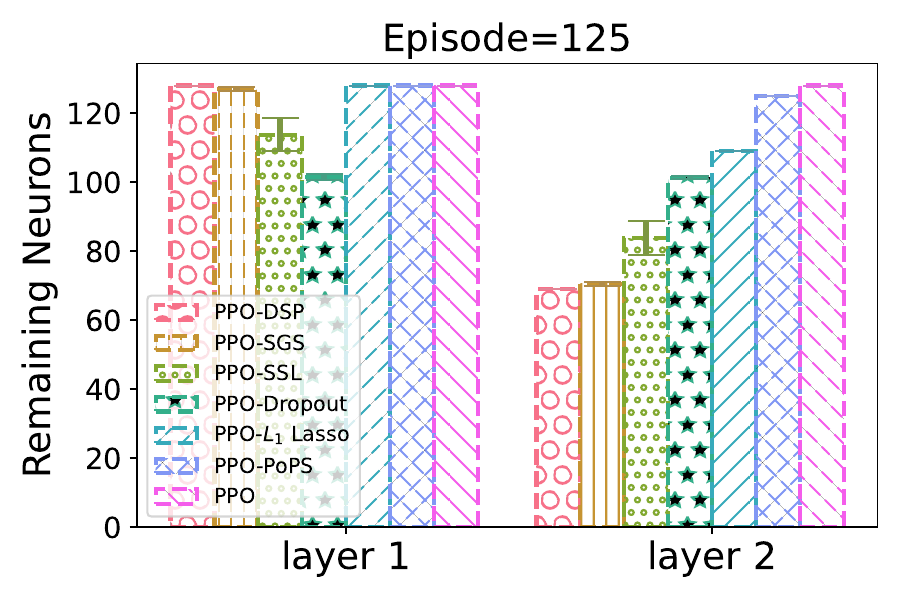}}
\subfloat[]{
\includegraphics[width=2.5in]{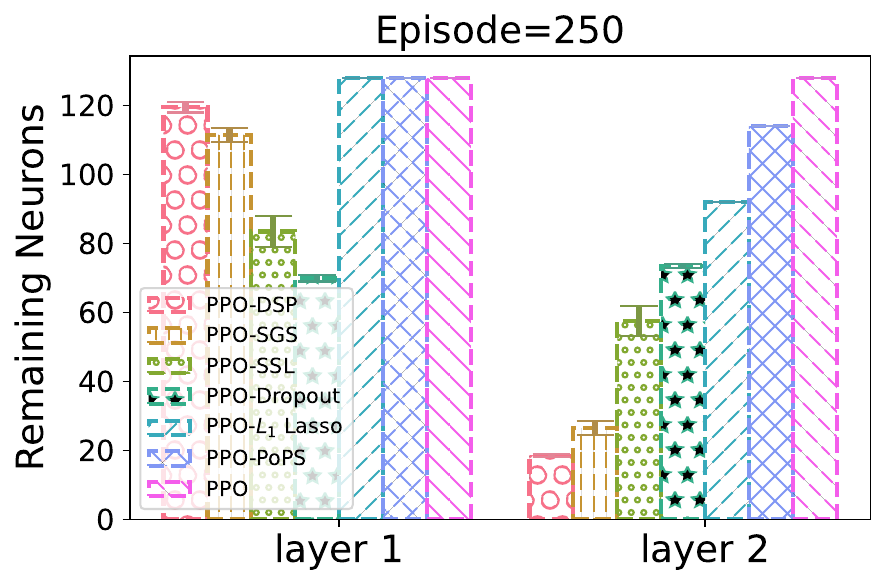}}

\subfloat[]{
\includegraphics[width=2.5in]{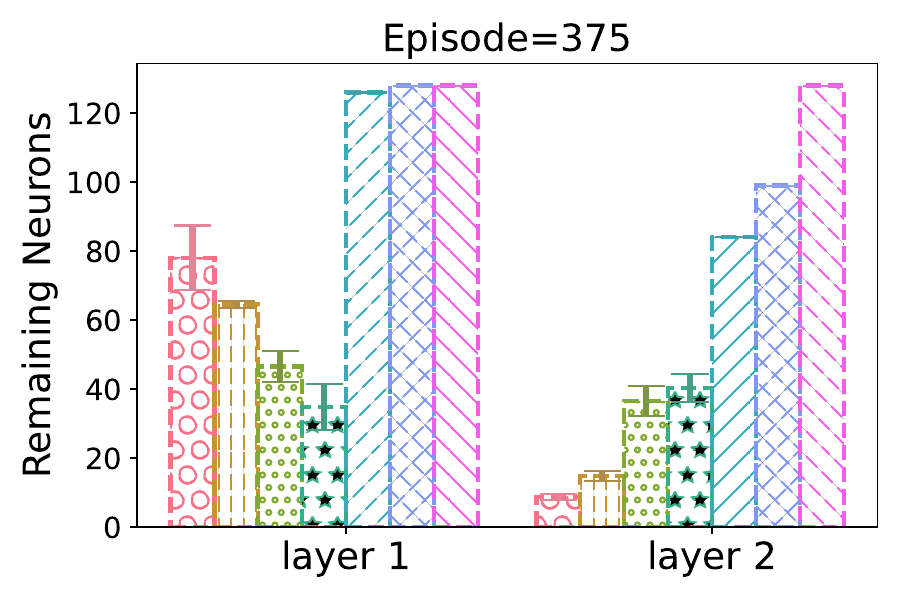}}
\subfloat[]{
\includegraphics[width=2.5in]{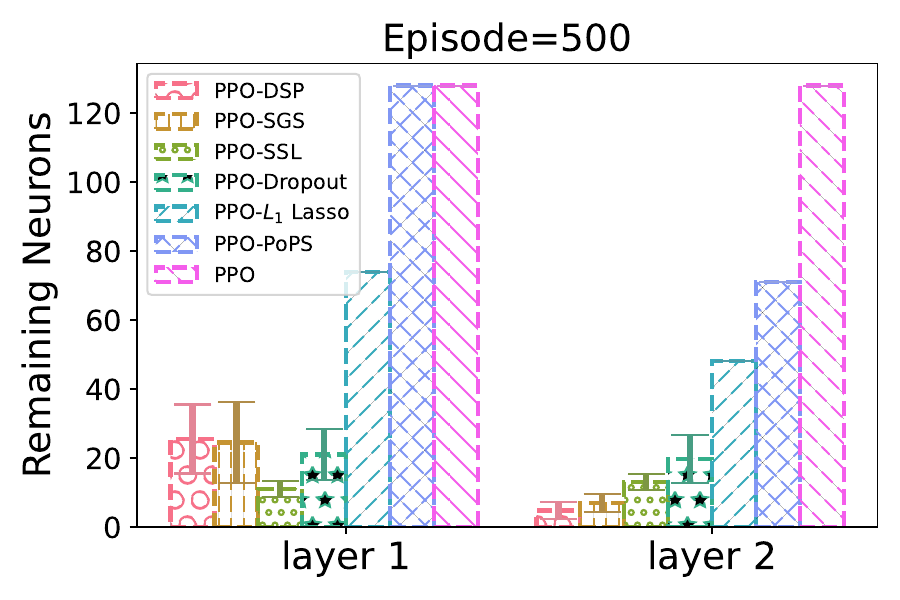}}
\end{center}
\caption{Analysis of visualization in the remaining weights per-layer of the actor networks under cartpole environment.}
\label{fig:E4-visualization}
\end{figure*}

\cref{fig:E4-visualization} shows how the number of neurons in each layer of the actor network changes with the episodes for different pruning methods. It can be seen that the proposed PPO-DSP method can achieve a higher compression rate than the baselines, especially in the second hidden layer, where the number of neurons is reduced by more than 97\%. This indicates that PPO-DSP can effectively prune the unimportant neurons and weights of the DRL model while maintaining high performance. In addition,~\cref{fig:E4-visualization} compares the proposed PPO-DSP method with other pruning methods, such as PPO-SSL, PPO-Dropout, PPO-L1 Lasso, PPO-PoPS, and PPO-SGS. It can be observed that PPO-DSP has a faster convergence rate and a lower variance than the other methods, which means that PPO-DSP can learn a stable and efficient policy with fewer neurons and weights. Moreover, PPO-DSP outperforms the other methods in terms of the final compression rate, which demonstrates the superiority of the dynamic structured pruning method based on a neuron-importance group sparse lasso.

Moreover,~\cref{fig:E4-visualization} also reveals the difference between the first and the second hidden layer in terms of the pruning effect. It can be seen that the second hidden layer has a higher pruning rate and a faster pruning speed than the first hidden layer for all the methods. This suggests that the second hidden layer may contain more redundant and unimportant neurons than the first hidden layer, and thus can be pruned more aggressively without affecting the performance. On the other hand, the first hidden layer may have more important neurons that are related to the input features and thus need to be preserved more carefully. The proposed method for the importance of neurons can screen out important neurons more effectively and prune redundant weights of the DRL network.

\section{Conclusion}\label{sec:Conclusion}
In this paper, we have proposed a novel dynamic structured pruning method for DRL models compression based on a neuron-importance group sparse lasso, which consists of two steps, i.e., training DRL models with a convex regularizer and removing unimportant neurons with the dynamic pruning threshold. To train the DRL model efficiently while focusing on a small number of important neurons, we use a neuron-importance group sparse regularizer which penalizes redundant groups of neurons and encourages the optimization process to aim for both high accuracy and a small number of important neurons. Additionally, we have developed a structured pruning strategy that dynamically determines the pruning threshold and gradually removes unimportant neurons using a binary mask. Our method not only removes redundant and unimportant neurons from the DRL model but also maintains high and robust performance.

Experiments on continuous and discrete DRL environments have demonstrated that our method outperforms existing pruning techniques both in terms of compression and in the ability to preserve the network’s performance. Moreover, the proposed method can largely remove the unnecessary neurons compared with the PPO-SSL while keeping the control performances. In detail, extensive experiments on CartPole-v1 and LunarLander-v2 have shown our method performs less than 8.2\% performance degradation and up to 93\% and 96\% neurons and weights reduction, respectively. Moreover, the proposed method can perform over the baselines 12.7\% and 23.2\% with the same parameters reduction in Walker2d-v3 and Hopper-v3.

In the future, optimizing neural networks and the selection of network parameter initialization will become more important, especially in manufacturing sectors using advanced robotics. For example, in automated product assembly lines where accuracy and speed are crucial, the balance between high-performance computing and network structure compression affects operational efficiency. Using hyperparameter optimization tools like KATIB is essential in these scenarios, which ensures a balance to ensure fast, real-time decision-making while reducing computational load.

\bibliographystyle{IEEEtran}
\bibliography{reference.bib}

\end{document}